%% file: main.tex
\documentclass[twocolumn]{bytedance}
\usepackage[toc,page,header]{appendix}

\usepackage{amsmath}
\usepackage{graphicx}
\usepackage{listings}
\usepackage{booktabs}
\usepackage{hyperref}
\hypersetup{colorlinks=true,breaklinks,linkcolor=red,citecolor=cyan}
\usepackage{url}
\usepackage{float}
\usepackage{amsfonts}
\usepackage{adjustbox}
\usepackage{xspace}
\usepackage{wrapfig}
\usepackage{amssymb}
\usepackage{color}
\usepackage{colortbl}
\usepackage[table]{xcolor}
\usepackage{caption}

\usepackage{algorithmic}
\usepackage[linesnumbered,ruled,vlined]{algorithm2e} %

\usepackage{multirow}
\usepackage[dvipsnames,table]{xcolor}
\usepackage{utfsym}
\usepackage{pifont}

\newcommand{\rev}[1]{#1}
\newenvironment{revblock}{\ignorespaces}{}
\newcommand{\del}[1]{}

\newenvironment{IEEEkeywords}%
{\par\noindent\small\textbf{Index Terms---}}%
{\par}

\title{
 Sa2VA: Marrying SAM2 with MLLM for Dense Grounded Understanding of Images and Videos
}

{
\small
\author[]{Haobo Yuan$^{1*}$}
\author[\dagger\ddagger]{Xiangtai Li$^{2*}$}
\author[]{Tao~Zhang$^{2,3*}$}
\author[]{Yueyi Sun$^{4}$}
\author[]{Zilong~Huang$^{2}$}
\author[]{Shilin~Xu$^{4}$}
\author[]{Shunping~Ji$^{3}$}
\author[]{Yunhai~Tong$^{4}$}
\author[\ddagger]{Lu Qi$^{3}$}
\author[]{Jiashi Feng$^{2}$}
\author[]{Ming-Hsuan Yang$^{1}$}
}
\affiliation[]{
{University of California, Merced$^{1}$} \quad {Bytedance Seed$^{2}$} \quad {Wuhan University$^{3}$} \quad {Peking University$^{4}$}\\
\vspace{3pt}
{$^\dag$Project lead} \quad {$^*$Equal technical contribution} \quad {$^\ddag$Corresponding authors} \\
\vspace{3pt}
{Code: \ \url{https://github.com/Bytedance/Sa2VA}} \\
\vspace{3pt}
{Model: \ \url{https://hf.co/collections/ByteDance/sa2va-model-zoo}}
}

\input{tex/0_abs.tex}

\date{\today}
\correspondence{Xiangtai Li at \email{xiangtai94@gmail.com}, Lu Qi at \email{luqi360@whu.edu.cn}}

\begin{document}
\maketitle

\input{tex/0_teaser}
\input{tex/0_keywords}
\input{tex/1_intro}

\input{tex/2_related_work}
\input{tex/3_method}
\input{tex/4_exp}
\input{tex/5_con}

\bibliographystyle{plainnat}
\bibliography{egbib}

\clearpage
\beginappendix
\setcounter{figure}{0}
\setcounter{table}{0}
\renewcommand{\thefigure}{A\arabic{figure}}
\renewcommand{\thetable}{A\arabic{table}}
\input{tex/6_appendix}

\end{document}

%% file: tex/0_abs.tex
\abstract{
This work presents Sa2VA, the first comprehensive, unified model for dense grounded understanding of both images and videos.
Unlike existing multi-modal large language models, which are often limited to specific modalities and tasks, Sa2VA supports a wide range of image and video tasks, including referring segmentation and conversation, with minimal single-stage instruction tuning.
Sa2VA combines SAM-2, a foundation video segmentation model, with MLLM, advanced vision-language models, and unifies text, image, and video into a shared LLM token space.
Using the LLM, Sa2VA generates instruction tokens that guide SAM-2 in producing precise masks, enabling a grounded, multi-modal understanding of both static and dynamic visual content.
Additionally, we introduce Ref-SAV, an auto-labeled dataset containing over 72k object expressions in complex video scenes, designed to boost model performance.
We also manually validate 2k video objects in the Ref-SAV datasets to benchmark referring video object segmentation in complex environments.
Experiments show that Sa2VA achieves strong performance across multiple tasks, particularly in referring video object segmentation, highlighting its potential for complex real-world applications.
In addition, Sa2VA can be easily extended into various MLLMs, including Qwen-VL and Intern-VL, which can be updated with rapid process in current open-sourced MLLMs.
Code and models have been provided to the community.
}

%% file: tex/0_teaser.tex
\begin{figure*}[t]
  \centering

  \includegraphics[width=.9\textwidth]{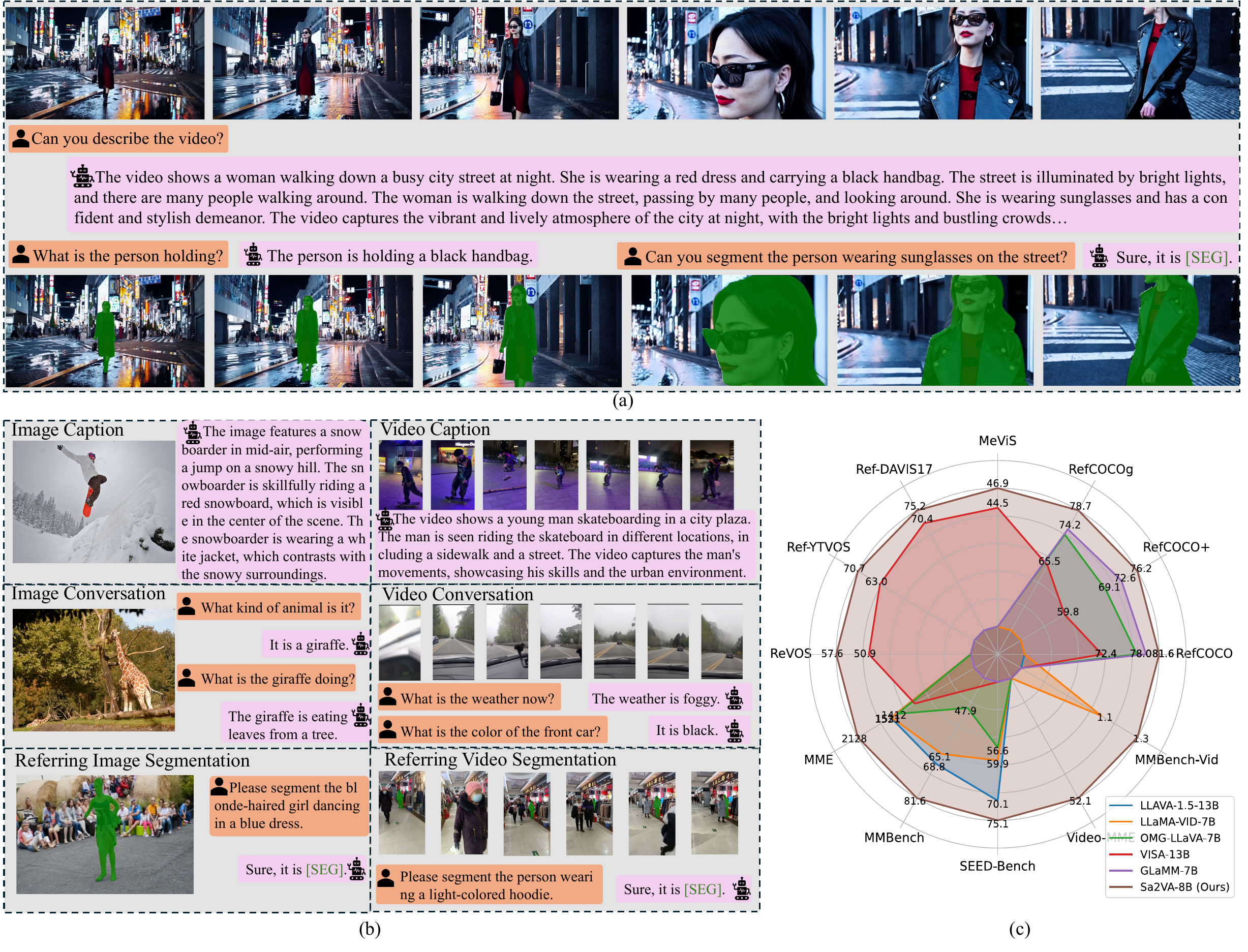}
      \vspace{-2mm}
  \caption{\textbf{Illustration of capabilities of our proposed Sa2VA.} (a). Given a video, Sa2VA can segment the referred object and understand the whole scene. (b). Sa2VA supports image chat, video chat, image referring segmentation, video referring segmentation, and grounded caption generation with single-stage instruction-tuning. (c). Sa2VA achieves strong results on multiple image, video referring segmentation, and chat benchmarks compared to existing MLLMs~\cite{hanoona2023GLaMM,OMGLLaVA,yan2024visa}.}
  \label{fig:teaser}
\end{figure*}

%% file: tex/0_keywords.tex

%% file: tex/1_intro.tex
\section{Introduction}
\label{sec:intro}

Multi-modal Large Language Models (MLLMs) have made significant progress, fueled by the rapid development of Large Language Models (LLMs)~\cite{llama,llama2, qwen2}. Numerous MLLMs have been applied to image- and video-level tasks such as visual question answering (VQA)~\cite{antol2015vqa,shao2023prompting}, narrative story generation~\cite{yang2024seedstory,he2023animateastory,zhou2024storydiffusion}, and interactive editing~\cite{wang2024emu3,kondratyuk2023videopoet,huang2024style}. One important direction is to understand video content in a fine-grained manner, including segmenting and tracking pixels with language descriptions, and performing VQA on visual prompts in the video. In particular, we aim to achieve promptable fine-grained analysis of video, enabling the user to be in the loop when playing the video in an interactive mode, as shown in Fig.~\ref{fig:teaser} (a). This capability will enable various applications, such as short video editing~\cite{ceylan2023pix2video,wang2023zero,liu2024video}, robot navigation~\cite{guan2022ga,guan2023vinet,griffin2020video}, and surveillance analysis~\cite{patil2021unified}.

However, neither state-of-the-art video perception models~\cite{cheng2024putting,OMGSeg,zhang2023dvis,ravi2024sam2} nor video MLLMs~\cite{zhang2023videollama,li2023llamavid,han2024free,liu2024llavanext,li2024aria} can achieve this.
The former are limited by constrained semantic concepts and lack open-ended capabilities, such as video question answering (Video-QA)~\cite{li2023llamavid,li2023videochat}. While the state-of-the-art SAM-2 model~\cite{ravi2024sam2} can perform promptable segmentation and tracking, it cannot handle text-aware tasks, such as understanding language expressions or video conversation.
On the other hand, video MLLMs can understand videos and perform Video-QA. For example, the latest LLaVA~\cite{li2024llava_onevision} can achieve strong results on Video-QA. However, it can neither carry out perception tasks nor understand visual prompts
Several works~\cite{OMGLLaVA, fei2024vitron,yan2024visa,wu2025grounded,rang2025eve} have explored the combination of perception models and MLLMs. However, existing works mainly focus on exploring image tasks~\cite{OMGLLaVA, fei2024vitron,yuan2023osprey} or trying to solve one specific video task~\cite{yan2024visa,munasinghe2024videoglamm,bai2024one,jie2024memory}. To the best of our knowledge, \textit{no work} has successfully integrated the strengths of both sides to build a unified model that supports both image and video inputs across a diverse range of tasks (Tab.~\ref{tab:ability_compare}).

Based on the above analysis, we set out to unify two visual foundation models, SAM-2~\cite{ravi2024sam2} and MLLMs~\cite{liu2023llava} into one framework. 
As such, our model not only inherits the spatial-temporal perception capabilities of SAM-2 and the open-ended abilities of MLLMs but also benefits from learning additional knowledge derived from the additional training data. 
To achieve this goal, three key challenges must be addressed: (1) \textit{Task formulation}: How to effectively formulate a range of tasks into a unified training setting, particularly for multi-modal inputs. 
(2) \textit{Performance balance}: How to resolve conflicts between tasks,  such as ensuring strong referring visual understanding abilities without sacrificing the language proficiency of MLLMs. 
Note that existing methods~\cite{OMGLLaVA, wu2024flmm} find that conversation tasks are drastically degraded when grounding tasks are performed. The models are thus degraded to specialists.
(3) \textit{Knowledge Inheritance}: How to leverage the pre-trained knowledge from SAM-2 and MLLMs to build a unified model that is both robust and flexible with different foundation models. We note that SAM-2 is trained on more than 1B masks, and most MLLMs are trained with massive instruction-following data pairs. Furthermore, both model families are evolving rapidly, and a flexible framework is crucial to more easily leverage their future progress.

\input{tables/t1_com}

In this work, we present Sa2VA, the first model that integrates SAM-2 with MLLMs, creating a unified, grounded understanding of both images and videos. 
We first formulate various tasks into a single-stage visual instruction tuning process, including referring segmentation, visual question answering (VQA), and grounded conversation generation (GCG) for both image and video. 
We leverage the token-length flexibility of LLMs, treating all input images, videos, and visual prompts as tokens without additional specific design.
Through joint training, we demonstrate that conflicts between grounding and conversation tasks can be effectively resolved.
Unlike existing approaches that use MLLMs as agents or connectors to call on visual experts, our model is trained end-to-end, showcasing model and dataset scalability.
Moreover, we adopt a decoupled design in which SAM-2's encoder and memory module are frozen, allowing us to retain its tracking capabilities. 
This design also makes our method a flexible framework, enabling our model to evolve with increasingly powerful MLLMs.
To be more specific, our framework is compatible with advanced MLLMs, such as InternVL~\cite{zhu2025internvl3} and Qwen3-VL~\cite{xu2025qwen3}.

In addition, we empirically observe that existing video segmentation datasets are limited to small-scale collections of short clips with a limited number of occlusions. Thus, to bridge the gaps between current datasets and the real-world requirements of interactive applications, we introduce Ref-SAV: a new, challenging reference video segmentation dataset curated through an automated pipeline based on the recent SA-V datasets~\cite{ravi2024sam2}. Finally, we benchmark several existing models on this dataset and find significant room for further exploration.

Sa2VA is co-trained with multiple datasets, including image and video modalities and various tasks. Without any bells and whistles, Sa2VA achieves strong performance on six referring image and video segmentation datasets while maintaining strong image and video QA capabilities compared to previous MLLMs, as shown in Fig.~\ref{fig:teaser}(c). On the proposed Ref-SAV dataset, our method achieves over 15\% performance gain over existing approaches under the zero-shot test setting, and obtains even stronger results using a large-scale Ref-SAV training set (about 37k videos). As shown in Fig.~\ref{fig:teaser}(c), our work builds a new strong baseline for a unified, dense, grounded understanding of images and video. Our main contributions are:

\begin{itemize}
    \item We develop Sa2VA, the first simple framework that combines SAM-2 and MLLM into one model. Sa2VA achieves strong spatial-temporal grounded understanding performance across various benchmarks. Sa2VA can be built upon with various modern MLLMs, including InternVL~\cite{zhu2025internvl3} and Qwen-VL~\cite{bai2025qwen2}.
    \item We introduce a challenging human-checked video benchmark, Ref-SAV. The benchmark introduces heavier occlusions, long text inputs, and motion blurs. 
    \item We develop a simple data annotation pipeline to build the Ref-SAV training dataset, where we find that the training dataset improves model performance on Ref-SAV.
    \item Extensive experiments show the effectiveness of Sa2VA on various open-sourced benchmarks (over 15 benchmarks, including image, video VQA, visual prompt, referring image/video segmentation). To the best of our knowledge, few existing MLLMs have shown comparable performance across such a diverse set of domains.
\end{itemize}

%% file: tables/t1_com.tex
\begin{table*}[t!]
    \centering
    \caption{\textbf{Capabilities of different representative models.} We compare Sa2VA against other representative works. As shown, while there are several groups of models that share similar capability sets (e.g., focusing on video conversation or dense grounding), no single existing method covers the full spectrum of tasks. In contrast, Sa2VA is designed to comprehensively support all listed modalities and tasks (also see Fig.~\ref{fig:teaser} (a) and (b)).}
    \vspace{-0.3mm}
    \resizebox{1.\textwidth}{!}{
    \begin{tabular}{c|ccc|cccc|cccc|c}
    \toprule[0.2em]
    \multirow{2}{*}{Method}  & \multicolumn{3}{c|}{Support Inputs} & \multicolumn{4}{c|}{Dense Grounding} &  \multicolumn{4}{c}{Conversation} & \multicolumn{1}{c}{End to End } \\
    ~  & \small Image & \small Video & \small Visual Prompts & \small RES & \small Ref-VOS & \small Inter-VOS & \small GCG & \small Image-Chat & \small Video-Chat & \small Video Caption & \small Interactive Caption & \small Training \\
    \midrule
    
    \multicolumn{13}{l}{\textit{Image-Chat Only Models}} \\
    \cmidrule(r){1-1}
    LLAVA~\cite{liu2023llava} & \textcolor{NavyBlue}{\usym{2713}}  &  \textcolor{BrickRed}{\usym{2717}}  & \textcolor{BrickRed}{\usym{2717}}  &  \textcolor{BrickRed}{\usym{2717}} & \textcolor{BrickRed}{\usym{2717}} & \textcolor{BrickRed}{\usym{2717}} & \textcolor{BrickRed}{\usym{2717}} & \textcolor{NavyBlue}{\usym{2713}} & \textcolor{BrickRed}{\usym{2717}} & \textcolor{BrickRed}{\usym{2717}} & \textcolor{BrickRed}{\usym{2717}} & \textcolor{NavyBlue}{\usym{2713}} \\
    
    \midrule
    \multicolumn{13}{l}{\textit{Image \& Video Chat Models}} \\
    \cmidrule(r){1-1}
    LLaVA-OneVision~\cite{li2024llava_onevision} & \textcolor{NavyBlue}{\usym{2713}}  &  \textcolor{NavyBlue}{\usym{2713}} & \textcolor{BrickRed}{\usym{2717}}  &  \textcolor{BrickRed}{\usym{2717}} & \textcolor{BrickRed}{\usym{2717}} & \textcolor{BrickRed}{\usym{2717}} & \textcolor{BrickRed}{\usym{2717}} & \textcolor{NavyBlue}{\usym{2713}} & \textcolor{NavyBlue}{\usym{2713}} & \textcolor{NavyBlue}{\usym{2713}} & \textcolor{BrickRed}{\usym{2717}} & \textcolor{NavyBlue}{\usym{2713}}  \\
    InternVL~\cite{chen2024far} & \textcolor{NavyBlue}{\usym{2713}}  &  \textcolor{NavyBlue}{\usym{2713}} & \textcolor{BrickRed}{\usym{2717}}  &  \textcolor{BrickRed}{\usym{2717}} & \textcolor{BrickRed}{\usym{2717}} & \textcolor{BrickRed}{\usym{2717}} & \textcolor{BrickRed}{\usym{2717}} & \textcolor{NavyBlue}{\usym{2713}} & \textcolor{NavyBlue}{\usym{2713}} & \textcolor{NavyBlue}{\usym{2713}} & \textcolor{BrickRed}{\usym{2717}} & \textcolor{NavyBlue}{\usym{2713}} \\
    LLaMA-VID~\cite{li2023llamavid} & \textcolor{NavyBlue}{\usym{2713}} & \textcolor{NavyBlue}{\usym{2713}} & \textcolor{BrickRed}{\usym{2717}} & \textcolor{BrickRed}{\usym{2717}} & \textcolor{BrickRed}{\usym{2717}} & \textcolor{BrickRed}{\usym{2717}} & \textcolor{BrickRed}{\usym{2717}} & \textcolor{NavyBlue}{\usym{2713}} & \textcolor{NavyBlue}{\usym{2713}} & \textcolor{NavyBlue}{\usym{2713}} & \textcolor{BrickRed}{\usym{2717}} & \textcolor{NavyBlue}{\usym{2713}}  \\
    P-LLaVA~\cite{xu2024pllava} & \textcolor{NavyBlue}{\usym{2713}} & \textcolor{NavyBlue}{\usym{2713}} & \textcolor{BrickRed}{\usym{2717}} & \textcolor{BrickRed}{\usym{2717}} & \textcolor{BrickRed}{\usym{2717}} & \textcolor{BrickRed}{\usym{2717}} & \textcolor{BrickRed}{\usym{2717}} & \textcolor{NavyBlue}{\usym{2713}} & \textcolor{NavyBlue}{\usym{2713}} & \textcolor{NavyBlue}{\usym{2713}} & \textcolor{BrickRed}{\usym{2717}} & \textcolor{NavyBlue}{\usym{2713}}  \\
    
    \midrule
    \multicolumn{13}{l}{\textit{Visual Prompt \& Dense Grounding Models}} \\
    \cmidrule(r){1-1}
    Osprey~\cite{yuan2023osprey} & \textcolor{NavyBlue}{\usym{2713}}  &  \textcolor{BrickRed}{\usym{2717}}  & \textcolor{NavyBlue}{\usym{2713}} &  \textcolor{BrickRed}{\usym{2717}} & \textcolor{BrickRed}{\usym{2717}} & \textcolor{BrickRed}{\usym{2717}} & \textcolor{BrickRed}{\usym{2717}} & \textcolor{NavyBlue}{\usym{2713}} & \textcolor{BrickRed}{\usym{2717}} & \textcolor{BrickRed}{\usym{2717}} & \textcolor{BrickRed}{\usym{2717}} & \textcolor{NavyBlue}{\usym{2713}} \\
    VIP-LLaVA~\cite{cai2024vipllava} & \textcolor{NavyBlue}{\usym{2713}}  &  \textcolor{BrickRed}{\usym{2717}}  & \textcolor{NavyBlue}{\usym{2713}} &  \textcolor{BrickRed}{\usym{2717}} & \textcolor{BrickRed}{\usym{2717}} & \textcolor{BrickRed}{\usym{2717}} & \textcolor{BrickRed}{\usym{2717}} & \textcolor{NavyBlue}{\usym{2713}} & \textcolor{BrickRed}{\usym{2717}} & \textcolor{BrickRed}{\usym{2717}} & \textcolor{BrickRed}{\usym{2717}} & \textcolor{NavyBlue}{\usym{2713}} \\
    GLAMM~\cite{hanoona2023GLaMM} & \textcolor{NavyBlue}{\usym{2713}}  &  \textcolor{BrickRed}{\usym{2717}}  & \textcolor{NavyBlue}{\usym{2713}} & \textcolor{NavyBlue}{\usym{2713}} & \textcolor{BrickRed}{\usym{2717}} & \textcolor{BrickRed}{\usym{2717}} &\textcolor{NavyBlue}{\usym{2713}} & \textcolor{BrickRed}{\usym{2717}} & \textcolor{BrickRed}{\usym{2717}} & \textcolor{BrickRed}{\usym{2717}} & \textcolor{BrickRed}{\usym{2717}} & \textcolor{NavyBlue}{\usym{2713}} \\
    LISA~\cite{lai2023lisa} & \textcolor{NavyBlue}{\usym{2713}}  &  \textcolor{BrickRed}{\usym{2717}}  & \textcolor{BrickRed}{\usym{2717}} &  \textcolor{NavyBlue}{\usym{2713}} & \textcolor{BrickRed}{\usym{2717}} & \textcolor{BrickRed}{\usym{2717}} & \textcolor{BrickRed}{\usym{2717}} & \textcolor{BrickRed}{\usym{2717}} & \textcolor{BrickRed}{\usym{2717}} & \textcolor{BrickRed}{\usym{2717}} & \textcolor{BrickRed}{\usym{2717}} & \textcolor{NavyBlue}{\usym{2713}} \\
    OMG-LLaVA~\cite{OMGLLaVA} & \textcolor{NavyBlue}{\usym{2713}} & \textcolor{NavyBlue}{\usym{2713}} & \textcolor{NavyBlue}{\usym{2713}} & \textcolor{NavyBlue}{\usym{2713}} &  \textcolor{NavyBlue}{\usym{2713}} & \textcolor{BrickRed}{\usym{2717}} & \textcolor{BrickRed}{\usym{2717}} & \textcolor{NavyBlue}{\usym{2713}} & \textcolor{NavyBlue}{\usym{2713}} & \textcolor{BrickRed}{\usym{2717}} & \textcolor{BrickRed}{\usym{2717}} & \textcolor{NavyBlue}{\usym{2713}}  \\

    \midrule
    \multicolumn{13}{l}{\textit{Dense Grounding Models}} \\
    \cmidrule(r){1-1}
    F-LLM~\cite{wu2024flmm}  & \textcolor{NavyBlue}{\usym{2713}} & \textcolor{BrickRed}{\usym{2717}} & \textcolor{BrickRed}{\usym{2717}} & \textcolor{NavyBlue}{\usym{2713}} & \textcolor{BrickRed}{\usym{2717}} & \textcolor{BrickRed}{\usym{2717}} & \textcolor{BrickRed}{\usym{2717}} & \textcolor{NavyBlue}{\usym{2713}} & \textcolor{BrickRed}{\usym{2717}} & \textcolor{BrickRed}{\usym{2717}} & \textcolor{BrickRed}{\usym{2717}} & \textcolor{BrickRed}{\usym{2717}}  \\
    VISA~\cite{yan2024visa} & \textcolor{NavyBlue}{\usym{2713}}  &  \textcolor{NavyBlue}{\usym{2713}}  & \textcolor{BrickRed}{\usym{2717}} &  \textcolor{NavyBlue}{\usym{2713}} & \textcolor{NavyBlue}{\usym{2713}} & \textcolor{BrickRed}{\usym{2717}} & \textcolor{BrickRed}{\usym{2717}} & \textcolor{BrickRed}{\usym{2717}} & \textcolor{BrickRed}{\usym{2717}} & \textcolor{BrickRed}{\usym{2717}} & \textcolor{BrickRed}{\usym{2717}} & \textcolor{BrickRed}{\usym{2717}} \\
    HyperSeg~\cite{wei2024hyperseg}&  \textcolor{NavyBlue}{\usym{2713}}  &  \textcolor{NavyBlue}{\usym{2713}}  & \textcolor{BrickRed}{\usym{2717}} &  \textcolor{NavyBlue}{\usym{2713}} & \textcolor{NavyBlue}{\usym{2713}} & \textcolor{BrickRed}{\usym{2717}} & \textcolor{BrickRed}{\usym{2717}} & \textcolor{NavyBlue}{\usym{2713}} & \textcolor{BrickRed}{\usym{2717}} & \textcolor{BrickRed}{\usym{2717}} & \textcolor{BrickRed}{\usym{2717}} & \textcolor{NavyBlue}{\usym{2713}}  \\
     InstructSeg~\cite{wei2024instructseg} & \textcolor{NavyBlue}{\usym{2713}}  &  \textcolor{NavyBlue}{\usym{2713}}  & \textcolor{BrickRed}{\usym{2717}} &  \textcolor{NavyBlue}{\usym{2713}} & \textcolor{NavyBlue}{\usym{2713}} & \textcolor{BrickRed}{\usym{2717}} & \textcolor{BrickRed}{\usym{2717}} & \textcolor{NavyBlue}{\usym{2713}} & \textcolor{BrickRed}{\usym{2717}} & \textcolor{BrickRed}{\usym{2717}} & \textcolor{BrickRed}{\usym{2717}} & \textcolor{NavyBlue}{\usym{2713}} \\

    \midrule
    Sa2VA (Ours) & \textcolor{NavyBlue}{\usym{2713}} & \textcolor{NavyBlue}{\usym{2713}} & \textcolor{NavyBlue}{\usym{2713}} & \textcolor{NavyBlue}{\usym{2713}} & \textcolor{NavyBlue}{\usym{2713}} & \textcolor{NavyBlue}{\usym{2713}} & \textcolor{NavyBlue}{\usym{2713}} & \textcolor{NavyBlue}{\usym{2713}} & \textcolor{NavyBlue}{\usym{2713}} & \textcolor{NavyBlue}{\usym{2713}} & \textcolor{NavyBlue}{\usym{2713}} & \textcolor{NavyBlue}{\usym{2713}} \\
    \bottomrule[0.1em]
    \end{tabular}
    }
    \label{tab:ability_compare}
\end{table*}

%% file: tex/2_related_work.tex
\section{Related Work}
\label{sec:related_work}

\noindent
\textbf{Multi-modal Large Language Models.} Earlier works~\cite{li2022blip,li2023blip,huang2020pixelbert} explore better multi-modal fusion methods and feature extractors and design various fusion architectures, particularly for vision language tasks. 
With the advances of LLMs~\cite{gpt3,llama,llama2}, multi-modal instruction tuning on LLMs~\cite{Qwen-VL,team2023gemini,liu2023llava,chen2023llava} have recently attracted much attention.
Notably, various benchmarks~\cite{hudson2019gqa,liu2023mmbench,li2023evaluating,fu2023mme,tong2024cambrian1} have been established to enhance model performance, reinforcing the notion that data is crucial in developing state-of-the-art MLLMs.
One representative work, LLaVA~\cite{liu2023llava}, considers the visual features as visual tokens. 
Furthermore, LLaVA provides a unified data format to consolidate extensive VQA tasks.
Several works~\cite{yuan2023osprey} subsequently explore stronger visual cues to improve the visual inputs of LLaVA. 
Meanwhile, numerous methods~\cite{zhang2025llavagrounding, zang2023contextual, ren2023pixellm, internlmxcomposer, internlmxcomposer2, internlmxcomposer2_4khd, lin2023videollava, zhang2023videollama, qi2024generalizable,huang2024reason3d,lai2023lisa, zhang2025pixel} incorporate additional components to adapt LLaVA for visual grounding, detection, segmentation, and video VQA.
Recently, a new trend has emerged to unify image, video, and multi-image analysis within a single framework~\cite{li2024llava_onevision,liu2024llavanext}. 
LLaVA-OneVision~\cite{li2024llava_onevision} designs a single model to handle four different input sources.
In visual perception, several works~\cite{UNINEXT,OMGSeg,ravi2024sam2, zhang2025dvis++} also explore multi-dataset and multi-task co-training. 
SAM-2~\cite{ravi2024sam2} proposes a unified model for joint image and video interactive segmentation.
Our model, Sa2VA, integrates SAM-2 into existing MLLMs to create a unified end-to-end model. 
It aims to combine image and video for dense, grounded understanding, encompassing tasks such as segmentation, chat, and captioning.

\noindent
\textbf{Referring Segmentation.} This task aims to output specific masks (image) or tracked masks (video) driven by language description. 
Earlier studies~\cite{yu2018mattnet, GRES, khoreva2019video, MeViS, wu2024lgvi, zhang2024psalm} examine various fusion modules to enhance performance.
Then, several stronger models~\cite{UNINEXT} adopt DETR-like methods~\cite{detr,zhu2020deformabledetr,ni2024context,chendeco, zhou2024improving} to achieve unified segmentation and tracking in video. 
Equipped with LLMs, several recent works~\cite{lai2023lisa, OMGLLaVA, xia2023gsva, qi2024generalizable,wei2024a,chen2024revisiting} have been developed to accomplish more complex referring tasks, including reasoning for referring or joint mask and caption generation.
For example, LISA~\cite{lai2023lisa} involves reasoning-based segmentation. 
Then, GLaMM~\cite{hanoona2023GLaMM} annotates a new dataset and introduces region-level captioning and segmentation tasks.
Meanwhile, several recent models exploit joint instruction tuning for referring segmentation and conversation~\cite{yan2024visa,zhang2024groundhog,chen2023llava}. 
Our method expands on these studies in the video domain by utilizing SAM-2~\cite{ravi2024sam2}, while maintaining superior performance in both image/video referring tasks and conversation tasks.

\begin{figure*}[t]
  \centering
  \includegraphics[width=1.\linewidth]{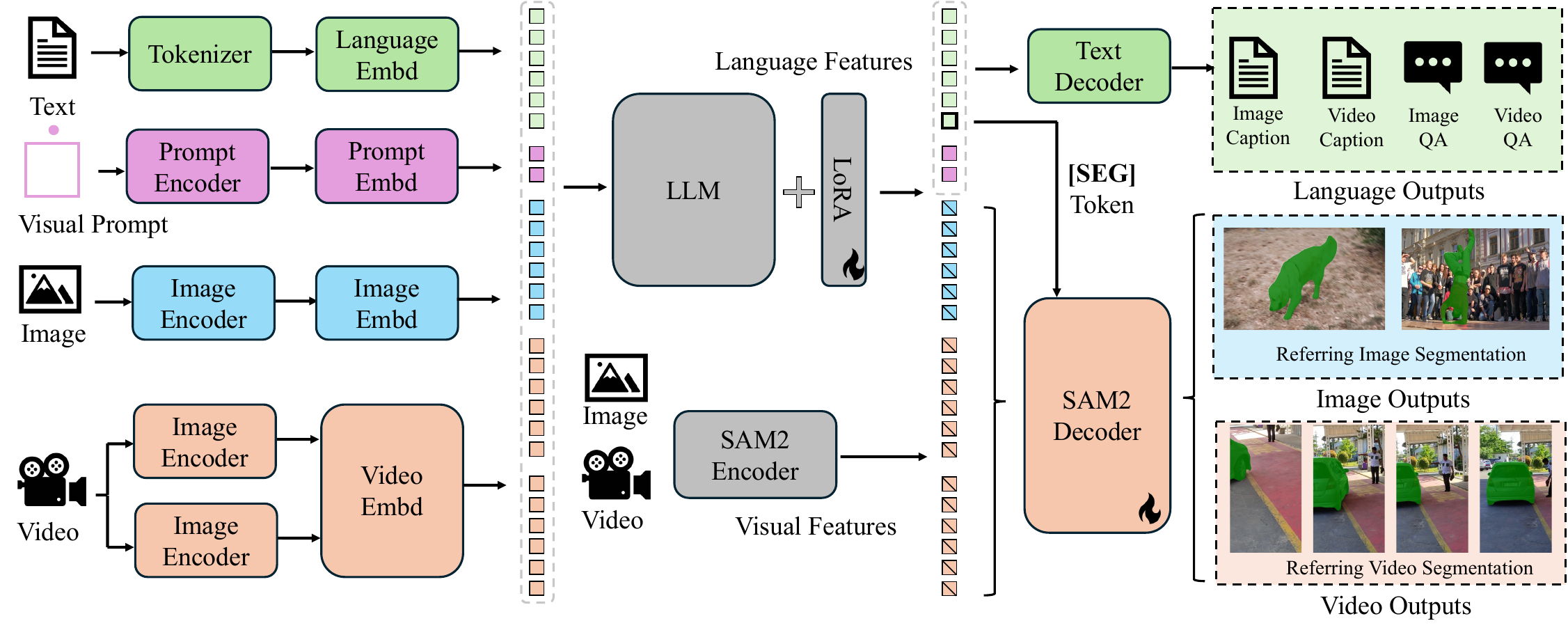}
  \caption{\textbf{Proposed Sa2VA model.} The model first encodes the input texts, visual prompts, images, and videos into token embeddings. These tokens are then processed through a large language model (LLM). The output text tokens are used to generate the ``[SEG]'' token and associated language outputs. The SAM-2 decoder receives the image and video features from the SAM-2 encoder, along with the ``[SEG]'' token, to generate corresponding image and video masks. Modules with a \textcolor{red}{fire} icon are trained during the single-stage instruction-tuning.}
  \label{fig:method}
\end{figure*}

\noindent
\textbf{Video Segmentation and Grounding.} Existing video segmentation methods~\cite{li2023tube,hwang2021video,zhu2022instance} mainly focus on segmenting and tracking pixels in closed sets, while 
a few approaches~\cite{guo2023openvis,zhou2023rethinking} explore open-vocabulary settings. 
However, concepts are still limited compared to the knowledge space of LLMs. 
For video grounding, a recent method~\cite{huang2024vtimellm} explores LLM in the joint understanding of video and audio.
On the other hand, VISA~\cite{yan2024visa} and VideoLISA~\cite{bai2024one} also explore referring video segmentation. 
However, they lack comprehensive training, which restricts their capability in other tasks and scaling up.
In contrast, Sa2VA enables fine-grained spatial-temporal modeling of both static (image) and dynamic (video) visual content, achieving strong performance across various tasks. 

%% file: tex/3_method.tex
\section{Method}
\label{sec:method}

\subsection{Unifying Multi-task Representations}
\label{sec:task_revisiting}

Developing a unified model to tackle various image and video understanding tasks is challenging due to their differences in spatial and temporal information.
To address this, we begin by re-examining the task formulations and proposing a unified representation, which lays the groundwork for the development of Sa2VA. 

\noindent
\textbf{Referring Image/Video Object Segmentation.} For image-referring segmentation, given input text tokens $T_{i} \in \mathbb{R}^{N\times D}$, the model processes input images $I_{i} \in \mathbb{R}^{H\times W\times 3}$ and produces corresponding binary masks $M_{o} \in \mathbb{R}^{H\times W}$ that align with the text description, where
$N$ and $D$ are the number and dimension of text tokens. 
For video object referring segmentation, the model takes input videos $V_{i} \in \mathbb{R}^{T\times H\times {W}\times 3}$ and outputs binary spatio-temporal masks (masklets) $M_{o} \in \mathbb{R}^{T \times H\times {W}}$ where $T$, $H$, and $W$ are video frame number, height, and width.

\noindent
\textbf{Image/Video Chat and Grounded Caption Generation.} For image and video chat tasks, given input text tokens $T_{i}$ and the corresponding images $I_{i}$ or videos $V_{i}$, the model produces the answer text $T_{o}$. For grounded caption generation tasks, the model simultaneously outputs the masks $M_{o}$ and the aligned text $T_{o}$.

\noindent
\textbf{Visual Prompt Understanding Tasks.} For visual prompt understanding tasks, in addition to text tokens $T_{i}$ and image $I_{i}$, the model also takes additional visual prompt tokens $VP_{i}$ (boxes or points on the image) as inputs, outputting corresponding masks $M_{o}$ and aligned text answers $T_{o}$.

\noindent
\textbf{Unified Task Representation.} Existing works address the above tasks through specific design models or partially unified models (different tasks utilize different model weights). In this work, we argue that all the aforementioned tasks can be unified into a single-stage instruction-tuning process, as we can leverage LLMs to manage various visual tokens. 
The overall process can be formulated as: 
\begin{equation}
    T_{o}, M_{o} = LLM(\{I_{i}, V_{i}, VP_{i}\}, T_{i}).
    \label{eq:unified}
\end{equation}
The input tokens vary for different tasks.
For chat-only tasks, the model only outputs text tokens $T_{o}$.
For referring segmentation tasks, the model outputs masks (image) or masklets (video) $M_{o}$. For grounded caption generation tasks, the model outputs both $T_{o}$ and $M_{o}$. 
Since the model primarily processes images/videos, text, and masks as inputs and outputs, it is reasonable to unify these tasks within a single framework and train the model end-to-end. This approach enables a single model to support all the aforementioned tasks effectively.

\subsection{Sa2VA Framework}
\label{sec:sa2va_framework_design}

The overall framework of Sa2VA is illustrated in Fig.~\ref{fig:method}. It consists of two components: the MLLM model and SAM-2.

\noindent
\textbf{Pre-trained MLLMs.} 
We adopt pre-trained MLLMs, which contain one visual encoder, one visual projection layer, and one LLM. 
The visual encoder takes input images, videos, and sub-images and outputs visual features.
The visual projection layer maps the features into tokens. 
These tokens, along with the input text tokens, serve as the input for LLMs, which generate text token predictions based on them. 
Note that we utilize pre-trained MLLMs, as seen in prior research~\cite{OMGLLaVA,hanoona2023GLaMM,lai2023lisa,yan2024visa}, to harness their powerful capabilities. 
For both image and video conversation tasks, we use the same pipeline as the original pretrained MLLMs (e.g., InternVL2~\cite{chen2024far}, Qwen-VL~\cite{Qwen-VL}).

\noindent
\textbf{Decoupled Design.} We append SAM-2~\cite{ravi2024sam2} alongside the pre-trained MLLMs.
Note that we do not take the SAM-2's output tokens (visual features or decoder outputs) into LLM. 
There are three reasons behind this design choice. 
First, we aim to simplify the combination without incurring additional computation costs.
Second, adding extra tokens will impede knowledge inheritance from MLLMs because MLLMs and SAM-2 are misaligned. 
Third, with this design, we can fully convert our work into a flexible framework that utilizes pre-trained, evolving MLLMs, as the MLLM community progresses rapidly.

\begin{algorithm}[!t]
\caption{Ref-VOS Inference Pipeline}\label{alg:refvos_inf}
\textbf{Input:} Video length $N$; Number of key frames $M$; Video frames $S_{N}$ ($X_1$, $X_2$, $X_3$,$\ldots$, $X_N$); Language description $T$;\\
\textbf{Output:} Sequence of masks $M_1$, $M_2$, $M_3$,$\ldots$, $M_N$;\\
\textbf{Run:} Sa2VA Model for Ref-VOS;\\
Extract key frames: $S_{M}$ $\gets$ $S_{N}$;\\
Visual embeddings: $E_v$ $\gets$ Encoder($S_{M}$);\\
Language embeddings: $E_l$ $\gets$ Encoder($T$);\\
Answers: $A$ $\gets$ LLM($\{E_v, E_l\}$);\\
Prompt embedding: $P_l$ $\gets$ Linear(Find($A$, '[SEG]'));\\
\For{$i = 1,2,\ldots,M$}{
    SAM-2 feature: $F_{i}$ $\gets$ Encoder($X_0$);\\
    Mask: $M_i$ $\gets$ Decoder($\{P_l, F_{i}\}$);\\
    Update Memory: $Mem$ $\gets$ Cross-Attention($\{Mem, M_i\}$);\\
}
\For{$i = M+1,M+2,\ldots,N$}{
    SAM-2 feature: $F_{i}$ $\gets$ Encoder($X_0$);\\
    Mask: $M_i$ $\gets$ Decoder($\{Mem, F_{i}\}$);\\
    Update Memory: $Mem$ $\gets$ Cross-Attention($\{Mem, M_i\}$);\\
}
\textbf{emit} $M_1$, $M_2$, $M_3$,$\ldots$, $M_N$;
\end{algorithm}

\noindent
\textbf{Tuning SAM-2 Decoder via ``[SEG]'' Tokens.}
Similar to previous works~\cite{lai2023lisa,yan2024visa} which utilize SAM, we employ the special token ``[SEG]'' to connect the decoder of SAM-2 with the MLLM.
The hidden states of the ``[SEG]'' token serve as a new kind of prompt, which are fed into SAM-2's Decoder, where they are decoded into segmentation masks. 
The hidden states of ``[SEG]'' can be viewed as a novel spatial-temporal prompt for SAM-2. SAM-2 segments the relevant object mask in images and videos based on this spatial-temporal prompt.
During training, the SAM-2 decoder can be adjusted to comprehend the spatial-temporal prompt, and gradients can be backpropagated through the ``[SEG]'' token to the MLLM, enabling it to gain knowledge from the training datasets.

\noindent
\textbf{Utilizing SAM-2 knowledge for mask tracking.} For Ref-VOS tasks, we develop a straightforward framework to attain strong results on public benchmarks. 
Specifically, we initially extract key frames (the first several frames) from the video. These key frames are then processed through MLLM. 
This MLLM then generates a ``[SEG]'' token as the SAM-2 prompt to create the masks for the key frames. 
Next, we utilize the memory encoded by the key frame features to generate the mask for the remaining frames. We present the default inference algorithm in Algorithm~\ref{alg:refvos_inf}.

\begin{revblock}
\noindent
\textbf{Benefits of a unified framework.} Compared with training separate task-specific models, a single shared set of weights provides four concrete benefits. (1) \emph{Positive cross-task transfer}: joint co-training across image/video~$\times$~chat/segmentation produces gains that cannot be recovered by single-task training alone (Tab.~\ref{tab:data_ablation}; e.g., removing image-segmentation data reduces RefCOCO performance from 77.4 to 20.2 cIoU). (2) \emph{Preserved conversational ability}: Sa2VA retains the chat capability of its base MLLM while achieving state-of-the-art grounding performance (Tab.~\ref{tab:image_benchmarks}). (3) \emph{Lower maintenance cost than per-dataset fine-tuning}: dataset-specific fine-tuning brings only marginal improvements (Tab.~\ref{tab:ft-ref-seg-results}) but requires maintaining a separate checkpoint for each dataset. (4) \emph{Deployment efficiency}: a single model handles all tasks without per-task switching, and the appended SAM-2 module adds only about 220M parameters while running at 39.5 FPS; thus, the overall cost is dominated by the shared MLLM (Tab.~\ref{tab:benchmark}). We provide further discussion of these benefits in the Appendix.
\end{revblock}

\subsection{Ref-SAV Dataset and Benchmark}
\label{sec:ref_sam_bench}
\noindent
\textbf{Data Annotation Pipeline.}
We create an automatic annotation pipeline to generate referring object text expressions for the SA-V dataset~\cite{ravi2024sam2}. 
As illustrated in Fig.~\ref{fig:method_annotation_pipeline}, the pipeline consists of three stages: 
\begin{figure}[t]
  \centering
\includegraphics[width=1.0\linewidth]{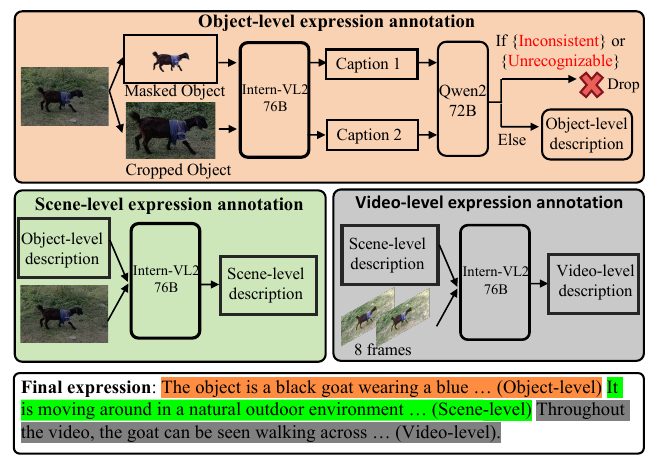}
  \caption{\textbf{Data annotation pipeline}. Our proposed automatic data annotation pipeline consists of three stages: object-level, scene-level, and video-level text expression annotation.}
  \label{fig:method_annotation_pipeline}
\end{figure}

(1) \textit{Object-level annotation}. We first select the frame with the largest object area from the video and mask out the non-object pixels. The cropped (without masking) and full (with masking) images are then fed separately into InternVL2-76B~\cite{chen2024far} to generate detailed descriptions. The descriptions are processed by Qwen2-72B~\cite{qwen2} for consistency checking, and conflicting descriptions are discarded. This method allows us to filter out error-prone cases.
(2) \textit{Scene-level annotation}. Both the image and the object-level description from the previous stage are sent into InternVL2-76B~\cite{chen2024far} to generate a detailed object description that includes relationships to the scene and surrounding objects.
(3) \textit{Video-level annotation}. We sample 8 frames uniformly from the video, applying yellow borders to emphasize the object in each frame. These frames, paired with the scene-level description, are processed by InternVL2-76B~\cite{chen2024far} to generate a video-level description that captures the object's movement and actions.

\noindent\textbf{Ref-SAV Dataset.} Using the above pipeline, we have automatically annotated detailed object expressions for the SA-V dataset, resulting in the creation of Ref-SAV. 
The training set of Ref-SAV contains 37,311 videos and 72,509 object expressions, which do not have any human labeling. 
For the Ref-SAV validation set, we select a subset of videos from the training set of the SA-V dataset to create the Ref-SAV evaluation benchmark, as the validation and test sets contain only a limited number of videos. 
These videos are completely separate from the Ref-SAV training dataset. 
The evaluation benchmark has two parts: 1) \textit{The long-expression set}, generated using the automatic annotation pipeline mentioned above and carefully filtered by human annotators. 
2) \textit{The short-expression set}, which is manually annotated. This evaluation benchmark includes 1,147 videos and 1,945 object expressions, comprising 1,694 long expressions and 251 short expressions. 
\rev{We note that the annotation pipeline uses a two-model agreement filter: captions generated by InternVL2-76B are cross-checked by Qwen2-72B. All reported test results are evaluated against human ground truth; moreover, the short-expression set is fully human-annotated, whereas the long-expression set is human-filtered. In the Appendix, we discuss the choice of annotation model, explain why it does not introduce an unfair family-internal advantage, and provide external validation of the automatically generated labels.}

\subsection{Sa2VA Training and Testing.}
\label{sec:model_training_testing}

\noindent
\textbf{One For All Co-Training.} Sa2VA is co-trained on multiple datasets. For VQA tasks, we utilize text regression loss $L_{text}$, similar to common MLLMs. For segmentation tasks $L_{mask}$, we employ pixel-wise cross-entropy loss $L_{CE}$ and dice loss $L_{DICE}$. It is important to note that there is no pre-training stage as in the base MLLMs~\cite{liu2023llava,chen2024far}; instead, we conduct supervised fine-tuning in a single-stage training with the following loss:
\begin{equation} 
    \mathcal{L}_{instruction} = \mathcal{L}_{text} + \mathcal{L}_{mask},\quad \mathcal{L}_{mask} = \mathcal{L}_{CE} + \mathcal{L}_{DICE}.\label{eq:sft} 
\end{equation}

\noindent
\textbf{One For All Testing.} All tasks can be encompassed within the Eq.~\ref{eq:unified} paradigm. During the inference stage, we encode the necessary task requirements, such as text prompts, visual prompts, image features, and video features, into tokens to input into the LLM. The output tokens from the LLM are then decoded into text responses (LLM prediction head), segmentation masks (SAM-2 decoder), and responses from the SAM-2 mask tracking module according to the task definition. 

%% file: tex/4_exp.tex
\section{Experiments}
\label{sec:exp}
\noindent
\textbf{Baseline.} We construct the baseline by combining the SOTA public MLLM model InternVL2~\cite{chen2024far} with SAM-2~\cite{ravi2024sam2}. Similar to previous works ~\cite{lai2023lisa, hanoona2023GLaMM, OMGLLaVA}, the segmentation mask is obtained by decoding the hidden state of the ``[SEG]'' token through SAM2's decoder. Inspired by Mask2Former-VIS~\cite{cheng2021mask2former}, an object shares the same ``[SEG]'' token throughout the video, enabling our model to handle both image and video referring segmentation tasks in a unified manner. In addition, we further use more advanced MLLM models, InternVL2.5~\cite{chen2024expanding}, to demonstrate the model scaling effect. \rev{Our paradigm treats the MLLM as a freely interchangeable plug-in, and it is our unified paradigm that endows an arbitrary conversational MLLM with strong referring segmentation while preserving its dialogue ability. Sa2VA thus does not stem from a stronger backbone, and even a relatively weak backbone (LLaVA-1.5-7B) plugged into our framework already surpasses the prior specialist (Appendix). Varying the backbone under our paradigm (Tab.~\ref{tab:exp_InternVL2_5}, Tab.~\ref{tab:exp_Sa2VA_Qwen}) yields consistent gains, confirming the improvement comes from the framework, not any particular backbone; the same modularity let us swap SAM-2 for SAM-3 (Appendix).}
\input{tables/t3_data_stat}
\input{tables/t_d1}
\input{tables/t_d2_stat}
\input{tables/t_impl}
\input{tables/t2_main}

\noindent
\textbf{Datasets and Metrics.} We use four types of datasets to train Sa2VA, which include image QA, video QA, image segmentation, and video segmentation datasets. As shown in Tab.~\ref{tab:train_datasets}, Sa2VA's training data includes approximately 1.1 million image-text or video-text pairs. Since InternVL2 has been trained with a large amount of image QA and video QA data, we only used 665K LLaVA 1.5~\cite{liu2023llavaplus} and 100K Video-ChatGPT~\cite{Maaz2023VideoChatGPT} data to prevent the MLLM from forgetting its image and video QA capabilities. 
We use 56K referring expression data~\cite{kazemzadeh2014referitgame, yu2016modeling} and 214K grounding conversation generation data~\cite{hanoona2023GLaMM} for image-level text-driven segmentation. For video-level referring expression segmentation, we used 5.8K existing referring VOS data from Ref-YouTubeVOS~\cite{seo2020urvos}, MeVIS~\cite{MeViS}, and ReVOS~\cite{yan2024visa}. Additionally, we used 37K long text referring VOS data (Ref-SAV), generated by our proposed automatic annotation pipeline, to enhance Sa2VA's understanding of long referring text and its object grounding capabilities for complex videos. For image referring segmentation, we adopt cIoU since it balances the large and small objects. For referring video object segmentation, we adopt J\&F. For image and video chat tasks, we follow previous works~\cite{liu2023llava,li2023llamavid} and report performance accordingly.

\noindent
\textbf{Implementation Details.} We adopt the xtuner~\cite{2023xtuner} framework for training. Our model inherits from MLLM models and does not require pre-training stages, which allows it to effectively leverage the information from the different base models. Unless specified otherwise, the following details describe the training of our default InternVL2 models. During the instruction tuning phase, we use the AdamW optimizer with an initial learning rate of 4e-5 and a weight decay of 0.05. We employ a 5\% warmup ratio and apply gradient clipping with max gradient norm as 1.0. For LoRA parameters, we set the rank (r) to 256.  The MLLM input image size is 448×448, and the SAM-2 input image size is 1024×1024. The maximum sequence length for the LLM is set to 8,192. We train for one epoch with a total batch size of 256 (16x2x8), achieved using a per-device batch size of 2 and 8 gradient accumulation steps. The training is conducted using bfloat16 (bf16) mixed precision on 16 NVIDIA H100 80GB GPUs. The instruction tuning stage lasts approximately 24 hours. The training strategy is listed in Tab.~\ref{tab:component_training_str} We adopt VLMEvalKit~\cite{duan2024vlmevalkit} for the evaluation of chat tasks, and we have provided our scripts for segmentation tasks. For Ref-SAV testing, we adopt the original open-sourced codebase~\cite{UNINEXT,wu2023uniref++,yan2024visa} and model weights to infer the video results. By default, we use single-stage instruction tuning and \textit{do not fine-tune on any specific task} to show the effectiveness of single-stage instruction tuning.

\input{tables/t5_image_level_bench_twoc}

\input{tables/t8_intern25}

\noindent
\textbf{Ref-SAV Dataset.} In Tab.~\ref{tab:Ref-SAV-benchmark-comparison}, we further compare the existing Ref-VOS benchmarks from five different aspects: short text, long text, large object motion, large camera motion, and heavy occlusion.
The previous benchmarks only contain partial aspects of these five challenging cases. On the other hand, our benchmark is built from SAM-2~\cite{ravi2024sam2} and labeled with long text, which is more challenging than previous benchmarks. This is why the previous Ref-VOS models cannot achieve good results on our benchmark.
Furthermore, as shown in Tab.~\ref{tab:tab_2_com_rebuttal}, our dataset has more detailed captions (Avg. Len. refers to the average length of the referring description) and many more masks (about 10x). The proposed data can lead to better instruction-following capability for complex descriptions.

\subsection{Main Results}
\label{sec:exp_main_results}
\noindent
\textbf{Comparison With SOTA MLLMs.}
As shown in Tab.~\ref{tab:exp_01_ref_comparision}, Sa2V-8B achieves cIoU scores of 81.9, 76.5, and 78.9 on RefCOCO, RefCOCO+, and RefCOCOg, respectively, surpassing GLaMM-7B by 2.4, 3.9, and 4.7 cIoU. 
Sa2VA performs favorably against the state-of-the-art methods on RefCOCO+ and RefCOCOg, significantly outperforming existing grounding MLLMs, including LISA, GLaMM, PixelLLM, PSALM, and OMG-LLaVA. 
Additionally, Sa2VA demonstrates strong conversational capabilities, achieving scores of 2229 (1651/578), 82.4, and 75.5 on MME, MMbench, and SEED-Bench, while existing grounding MLLMs perform poorly in conversation due to catastrophic forgetting.
Sa2VA achieves performance comparable to that of InternVL2 in image QA benchmarks, indicating that Sa2VA largely maintains the chat performance of the base MLLM.

Sa2VA also performs well on video benchmarks. 
It achieves scores of 46.9, 75.2, and 57.6 J\&F on MeVIS, Ref-DAVIS17, and ReVOS, respectively, surpassing the previous SOTA VISA-13B by 2.4, 4.8, and 6.7 J\&F. Additionally, Sa2VA-8B earns a score of 1.34 on the video QA benchmark MMBench-Video, outperforming InternVL2-8B's 1.28. 
Sa2VA does not achieve dominant results on video chat tasks, as it has both video segmentation and understanding capabilities, which current methods are not equipped to handle simultaneously.
The main experimental results indicate that our Sa2VA is a versatile and powerful MLLM.

\noindent
\textbf{Detailed Results on Image Benchmarks.} As indicated by~\cite{OMGLLaVA,lai2023lisa,wu2024flmm}, performing instruction tuning negatively impacts performance drastically on chat tasks. 
We utilize co-training to address this issue. In Tab.~\ref{tab:image_benchmarks}, we also present additional results on image chat datasets. Our model maintains good results across multiple image chat benchmarks while achieving strong referring segmentation performance, demonstrating its versatility.

\noindent
\textbf{Results on Ref-SAV validation set.} In Tab.~\ref{tab:ref-sam-v-results}, we benchmark several state-of-the-art Ref-VOS models using our proposed Ref-SAV benchmark. We find that even the foundation model~\cite{UNINEXT} and the recent video MLLM model~\cite{yan2024visa} fail to achieve strong results on our benchmarks. This is due to our benchmark featuring more occlusions, longer text descriptions, and diverse annotations from SAM-2~\cite{ravi2024sam2}. 
Conversely, our method, Sa2VA, whether with or without the Ref-SAV training set, can yield strong results. Our method can be further enhanced with our training set, indicating that our Ref-SAV training set serves as a valuable supplement to the video understanding community.

\input{tables/t7_cotrain}
\input{tables/t4_refsav_results}
\input{tables/t6_finetune}

\noindent
\textbf{Comparison with specific fine-tuned methods.} The original Sa2VA setting is more challenging because it requires the use of the same model weights to complete all tasks. For fair comparison, similar to prior works~\cite{hanoona2023GLaMM, OMGLLaVA}, we evaluate with MLLMs on specific fine-tuned datasets. 
As shown in Tab.~\ref{tab:ft-ref-seg-results}, we find 1.5\% improvement across three datasets compared to the co-training model, with Sa2VA performing best among recent methods. 
From another perspective, fine-tuning on a single dataset does not yield much benefit (0.4 on RefCOCO). However, this requires fine-tuning each dataset and producing a new model. Therefore, we advocate for more general models like Sa2VA for their versatility and convenience.

\noindent
\textbf{InternVL-2.5 Results.}
We further conduct ablation experiments to evaluate the scalability of Sa2VA with stronger MLLMs and additional datasets. As shown in Tab.~\ref{tab:exp_InternVL2_5}, upgrading the MLLM from InternVL2~\cite{chen2024far} to InternVL2.5~\cite{chen2024expanding} consistently improves performance across image, video, and multimodal benchmarks.

\subsection{Ablation Studies}
\label{sec:ablation_analysis}
\noindent
\textbf{Effectiveness of Joint Co-training.}
The performance of Sa2VA on image and video QA and segmentation benchmarks can be attributed to unified instruction tuning and joint co-training. 
We perform ablation studies on the training datasets, with the results presented in Tab.~\ref{tab:data_ablation}. 
When trained without image QA datasets, Sa2VA's scores on MME and MMBench drop by 129 and 4.9, respectively. 
Without image segmentation datasets, Sa2VA achieves only 20.2, 20.6, and 23.2 cIoU on RefCOCO, RefCOCO+, and RefCOCOg, while performance on the video segmentation benchmark drops significantly. 
Sa2VA on MMBench-Video decreases by 34\% when trained without video QA datasets. 
When training without video segmentation datasets, Sa2VA's performance on MeVIS and Ref-DAVIS17 drops by 6.4 and 3.3 J\&F, respectively. The ablation results indicate that joint co-training with these four types of datasets is critical to Sa2VA's performance across the various tasks.
In Tab.~\ref{tab:scale_data}, adding 3M image-QA samples from Infinity-MM~\cite{gu2024infinity} yields a notable +2.1 gain on MMBench, with negligible changes in segmentation accuracy. When incorporating our Ref-SAV dataset, Sa2VA-1B achieves a +1.7 J\&F improvement on MeViS while maintaining image-level performance. These results demonstrate that Sa2VA benefits synergistically from both model scaling and data expansion, indicating substantial headroom for further improvement.

\input{tables/t9_seg_token}
\input{tables/t10_dataset_ablation}

\input{tables/qwen_table}

\input{tables/t_a1_comp}
\input{tables/t_a2_comp_expert}
\input{tables/t_a3_refsav_training_effect}
\input{tables/t_a4_region_caption}

\input{tables/t_ana_1}

\noindent
\textbf{Ablation study on the Segmentation Token Design.} 
Tab.~\ref{tab:seg_token} shows the results of our ablation on the segmentation token design. We analyze two primary alternative strategies based on whether to use the same special token or different special tokens for different frames. First, we test the \textbf{repetitive} strategy, which requires the LLM to output the same ``[SEG]'' token $N$ times for $N$ frames. This approach slightly reduces video segmentation performance. Second, we analyze using \textbf{unique}, frame-specific tokens (e.g., ``[SEG 1]'' for frame 1, ``[SEG 2]'' for frame 2, and so on). This design caused a larger performance drop. The reason is that it prevents knowledge sharing with the image segmentation task. The model is typically pre-trained to associate a single, generic ``[SEG]'' token with the concept of segmentation. By introducing new tokens (``[SEG 1]'', ..., ``[SEG n]''), there is a mismatch between images and videos that blocks this knowledge transfer, forcing the model to learn the meaning of each new segmentation token independently. \rev{In other words, carrying richer or more specialized information through multiple dedicated tokens is a promising direction. However, it appears to require dedicated pre-training to re-establish the association between these tokens and segmentation outputs. This is beyond our single-stage instruction-tuning budget and is therefore left to future work.}

\subsection{More Comparison Results}

\noindent
\textbf{Comparison with recent video MLLM models on referral video segmentation} 
Recently, several works have also combined MLLMs into video referring segmentation tasks. As shown in Tab.~\ref{tab:video_mllm_models}, we list and compare our models against several recent methods on three datasets. These competitors include VideoGLaMM~\cite{munasinghe2024videoglamm}, VideoLISA~\cite{bai2024one}, HyperSeg~\cite{wei2024hyperseg}, InstructSeg~\cite{wei2024instructseg}, and etc. We also include the combined strong baseline (SAM2~\cite{ravi2024sam2} + GlaMM~\cite{hanoona2023GLaMM}) reported in~\cite{munasinghe2024videoglamm}. 
As shown in Tab.~\ref{tab:video_mllm_models}, our Sa2VA-8B model achieves strong performance across all three datasets. Our Sa2VA-4B model also demonstrates strong competitive performance, achieving the highest score on MeViS, while maintaining the ability for image/video chat, GCG, and visual prompt understanding. 
At the same time, other referring video segmentation methods only focus on this task, and due to the lack of training data for other tasks, the obtained model will perform poorly on other tasks such as video or image chat.

\noindent
\textbf{Comparison with vision expert models.} As shown in Tab.~\ref{tab:vision_expert_model}, we compare our model with recent vision expert models designed specifically for referring segmentation. Although these specialist models are typically more lightweight, our general-purpose 26B model (Sa2VA-26B) still achieves the strongest results, outperforming them across all five datasets. This is significant, as these expert models lack conversational capabilities and are mostly limited to a single modality (video or image), whereas our model is a generalist.

\noindent
\textbf{Effectiveness of training dataset on more methods.} We further demonstrate the effectiveness of our proposed Ref-SAV training dataset by evaluating a representative method, UniRef++~\cite{wu2023uniref++}. As shown in Tab.~\ref{tab:effectiveness_on_more_baselines}, we compare two distinct settings to isolate the impact of our data. The zero-shot setting involves directly testing the pre-trained UniRef++ model on the Ref-SAV validation dataset without any new training. In contrast, the fine-tuning setting first trains the model on our proposed Ref-SAV training dataset before evaluation on the same validation set. The results clearly show that the model fine-tuned on our dataset achieves a significant improvement across all metrics. Most notably, the Overall J\&F score increases substantially from 10.5 to 14.6. This considerable gain, which is consistent across both ``Long'' (12.5 to 17.2 J\&F) and ``Short'' (8.6 to 12.0 J\&F) validation splits, indicates that our automatic data engine has great potential for boosting the performance of Ref-VOS models.

\noindent
\textbf{Results on visual prompts understanding tasks.} We also report on visual prompt understanding tasks, following previous works~\cite{yuan2023osprey,OMGLLaVA}. Specifically, we evaluate region caption performance on the RefCOCOg dataset. As shown in Tab.~\ref{tab:effectiveness_visual_prompt_tasks}, our method, Sa2VA-4B, also achieves the best results among recent visual prompt understanding models. It scores 17.3 on the METEOR metric, significantly outperforming the next-best method, Osprey~\cite{yuan2023osprey}, which scored 16.6. This indicates that Sa2VA can also generate strong region-aware captions.

\subsection{Discussions}
\noindent
\textbf{Sa2VA with recent MLLM models.} Table \ref{tab:exp_Sa2VA_Qwen} presents a comprehensive evaluation of our Sa2VA framework integrated with different recent base Multimodal Large Language Models (MLLMs), including Qwen2.5-VL~\cite{bai2025qwen2}, Qwen3-VL~\cite{xu2025qwen3}, and InternVL-3~\cite{zhu2025internvl3}. The results clearly demonstrate that the choice of the base MLLM significantly influences performance, and this effect varies across different tasks. For instance, while the Sa2VA-InternVL3-14B model achieves the strongest overall performance, particularly on image and video segmentation benchmarks like RefCOCO (83.6) and MeViS (59.2), other models exhibit competitive strengths in specific domains. Notably, the Sa2VA-Qwen3VL-4B model records the highest score on MMBench (86.3) and shows highly competitive results on other image chat datasets. This variation highlights that different base MLLM architectures possess distinct capabilities, making model selection a critical factor for task-specific optimization. To facilitate further research and allow the community to build upon these findings, we have publicly released all trained model checkpoints.

\noindent
\textbf{Discussion on the inference speed.} The computational overhead in Sa2VA is dominated by the MLLM component due to its auto-regressive nature, where each inference step generates tokens sequentially. In contrast, the primary additional module, SAM-2, is relatively lightweight at 220M parameters (compared to the MLLMs, which range from 1B to 26B) and efficient, achieving 39.5 FPS on vision tasks with only a single forward pass per image. While the variable output length of the MLLM makes precise latency estimation challenging, to provide a more concrete comparison, we also construct a controlled benchmark with fixed frame and language lengths (``Please segment the man.'' as the text prompt with 5 frames and totaling 1280 tokens) to estimate computational cost more systematically. The results of this benchmark are detailed in Table~\ref{tab:benchmark}. The results show inference times of 0.123s for Sa2VA-1B, 0.282s for Sa2VA-4B, 0.201s for Sa2VA-8B, and 0.463s for Sa2VA-26B. We note that inference speed is largely dependent on optimization, and the discrepancy between the 4B model (0.282s) and the 8B model (0.201s) may be due to the different base LLMs (Qwen vs. InternLM) employed. Nonetheless, these results confirm that the majority of the model’s computational cost stems from the auto-regressive MLLM component rather than the efficient, single-pass SAM-2 module.

\noindent
\textbf{Inference Strategy.} We also analyzed the impact of our keyframe sampling strategy. Our default approach of using the first five frames was chosen for simplicity and consistency, but we investigated whether this method limits the model's ability to capture long-range temporal dependencies or objects that appear later in the video. To test this, we performed an ablation study on the MeVIS dataset using the Sa2VA-8B model. We compared the performance of selecting the first 1, 3, 4, and 5 consecutive frames against a uniform sampling strategy that selects 5 frames spread across the entire video. The results of this comparison are detailed in Table~\ref{tab:sampling_strategy}. As the table shows, performance generally improves as the number of consecutive frames increases, from 55.1 J\&F for a single frame to 59.5 J\&F for four frames. However, the ``First 5 Frames'' strategy (58.9 J\&F) is significantly outperformed by the ``Uniform 5 Frames'' strategy, which achieves 62.9 J\&F. Notably, both 5-frame methods have an identical inference time (0.207s) and process the same number of image tokens (1280). This finding suggests that more sophisticated, interval-based sampling strategies improve performance. However, we would like to keep Sa2VA as simple as possible and leave the design of advanced sampling techniques to future work.

\subsection{Visualizations}
\label{sec:visualization_results}
\begin{figure}[!t]
  \centering
  \includegraphics[width=1.0\linewidth]{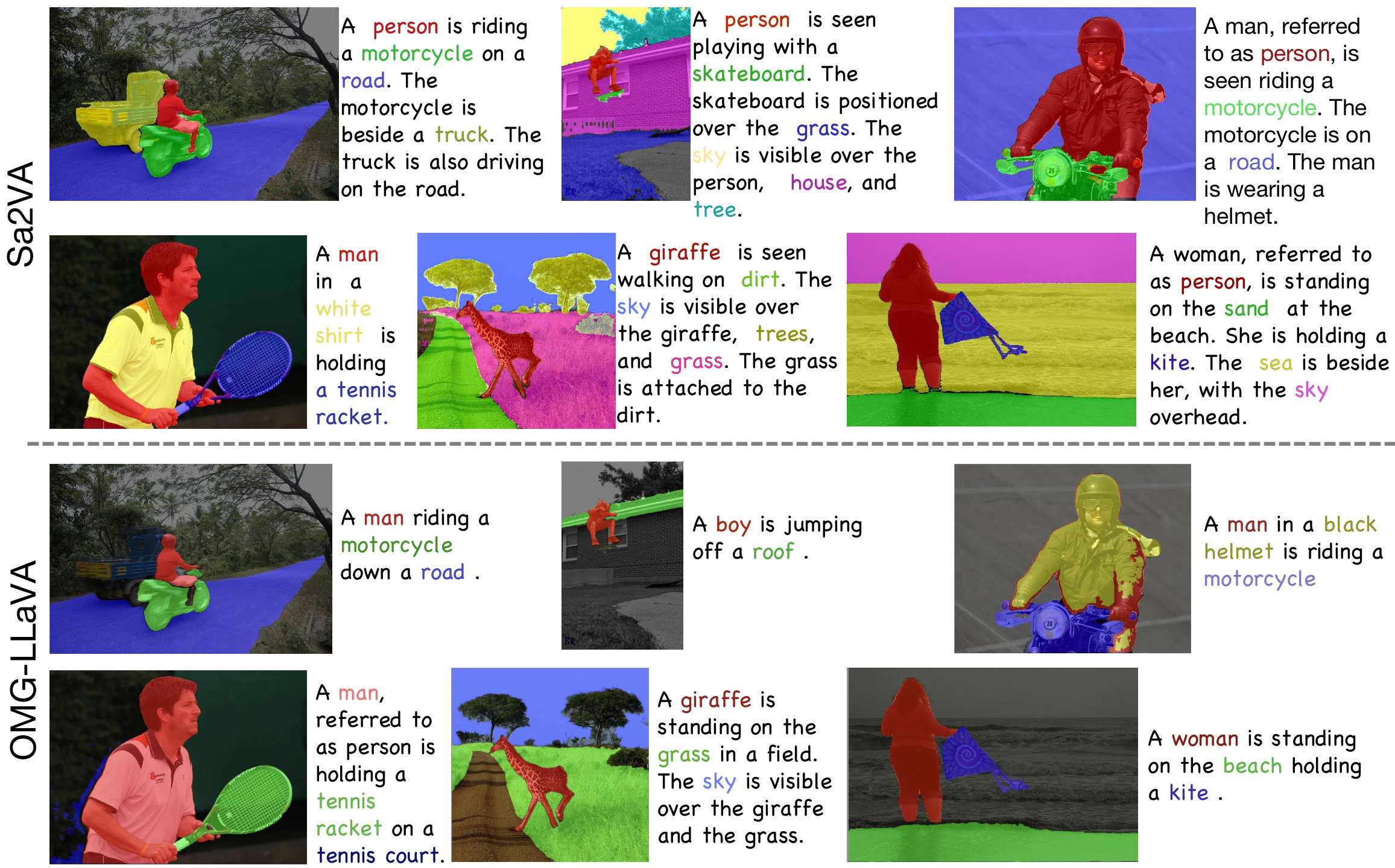}
  \caption{Visualization results on GCG tasks. Top: our method. Bottom: OMG-LLaVA~\cite{OMGLLaVA}. Note that our method has stronger, fine-grained grounding and text alignment than OMG-LLaVA~\cite{OMGLLaVA}, a strong previous baseline.}
  \label{fig:gcg_demo}
\end{figure}

\begin{figure}[!t]
  \centering
  \includegraphics[width=0.85\linewidth]{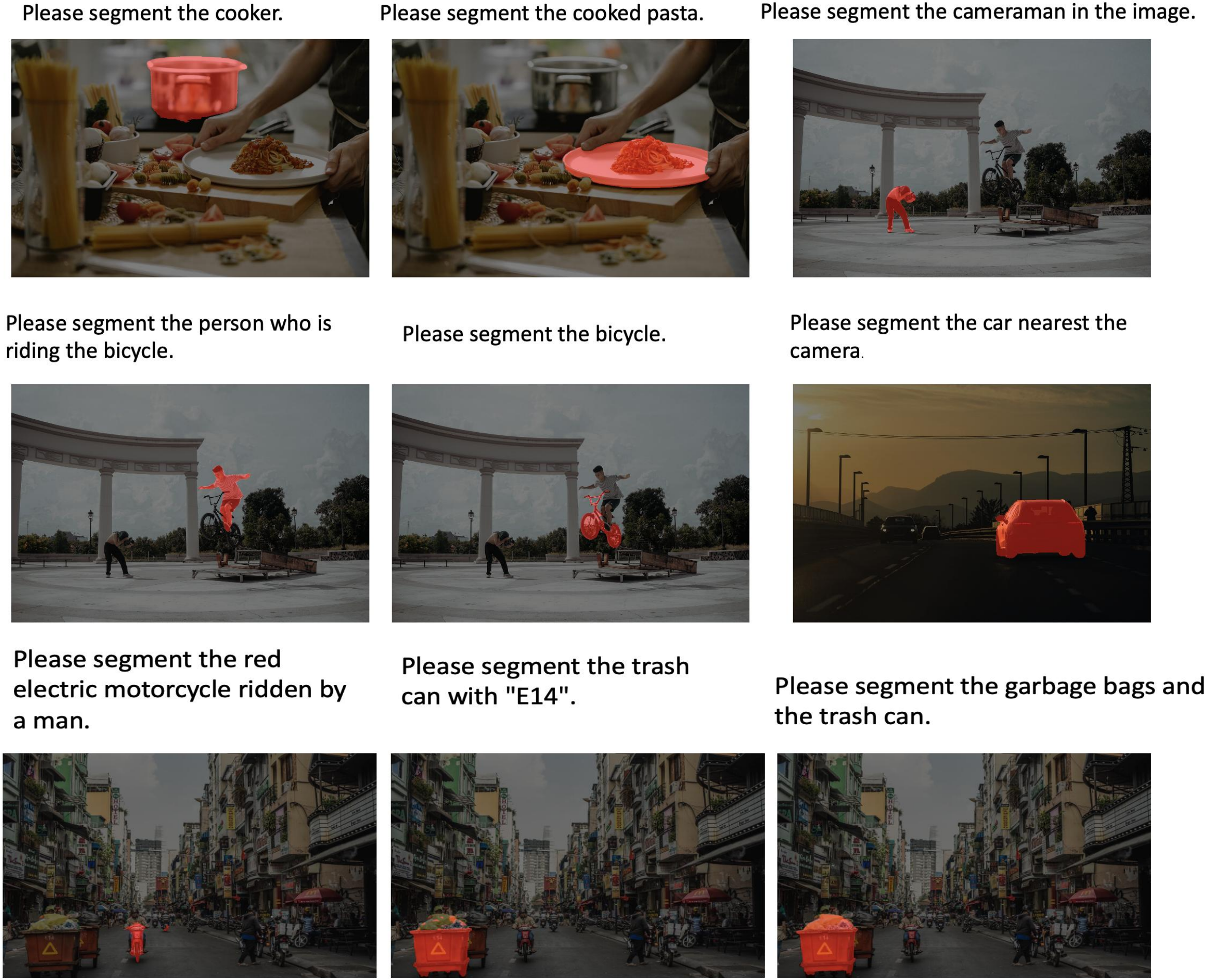}
  \caption{Visualization results on the image referring segmentation.}
  \vspace{-4mm}
  \label{fig:res_demo}
\end{figure}

\noindent
\textbf{Results on image referring segmentation.}
As shown in Fig.~\ref{fig:res_demo}, we visualize the image referring segmentation task. With different language descriptions, our Sa2VA can segment different masks with the clues from the description.
\rev{Additional qualitative results on video referring segmentation, visual-prompt understanding, and failure cases are provided in the Appendix.}

\noindent
\textbf{Results on GCG task.} 
In Fig.~\ref{fig:gcg_demo}, we further show several visual examples in grounded caption generation tasks. Compared with the previous SOTA model, OMG-LLaVA, our model achieves better results in both mask quality and text-mask alignments. The former indicates better-aligned segmentation outputs, while the latter shows the latter indicates good text-to-region alignments.

\input{tables/t_ana_2}

\noindent
\textbf{Video demo.}
Since dynamic visual content such as videos cannot be effectively presented within the static format of the manuscript, we have provided additional video demos in the supplementary material. These videos are intended to offer a clearer and more intuitive understanding of our method’s performance. Please refer to the ZIP file for the demos.

\section{Failure Cases and Future Work}
\label{sec:failure_cases_and_future_work}

\noindent
\textbf{Failure cases of Sa2VA and Future work.} \rev{Although Sa2VA achieves strong results across various referring segmentation datasets, there are still several failure cases that need to be explored.}
We identify two shortcomings of Sa2VA in our future work. One is a long video with hard, distinguished referring examples (Appendix). This is because our Sa2VA works in online mode without knowing the entire video content to align the long, complex text. \rev{Tab.~\ref{tab:sampling_strategy} shows that uniformly sampling 5 frames substantially outperforms using the first 5 frames ($62.9$ vs.\ $58.9$ J\&F on MeViS val\_u) at the same inference cost, indicating that broader temporal coverage is beneficial. We therefore view interval-based or adaptive temporal sampling, together with stronger long-range memory, as a principled remedy for long-video alignment.}
\rev{The other open challenge is how to improve VQA performance on both images and videos without degrading referring segmentation. As shown in the ablation table in the main paper, increasing the scale of VQA data leads to a performance drop on referring segmentation tasks. Thus, how to balance these two types of supervision while preserving grounded prediction knowledge remains an unsolved problem.}

\begin{revblock}
\noindent
\noindent
\textbf{Hypothesis on the VQA--grounding conflict.} We hypothesize that this residual imbalance reflects a \emph{representation--objective conflict} rather than a capacity limitation. VQA is optimized with a next-token language-modeling loss, which favors holistic, sequence-level semantic representations. In contrast, referring segmentation requires the hidden state of the ``[SEG]'' token to preserve fine-grained, spatially discriminative information for the SAM-2 decoder. As the amount of VQA data increases, the LLM representations may drift toward a global-semantic attractor, thereby weakening the spatial signal carried by ``[SEG]''. We further conjecture that visual generative objectives, such as image/video synthesis or masked reconstruction, may interfere \emph{less} with dense grounding because they also require spatially grounded representations, and may even serve as a mutual regularizer. We provide extended analysis and discuss possible mitigations in the Appendix.

\noindent
\textbf{Reinforcement learning as a future direction.} The language-modeling objective used in this work establishes a strong single-stage baseline. A natural next step is reinforcement learning with \emph{verifiable rewards}, such as mask IoU or boundary F-score, optionally combined with text-side preference signals as in RLHF/GRPO. Such task-specific rewards could inject grounding supervision only where it is needed, thereby mitigating the VQA--grounding interference discussed above and improving robustness on challenging, distinguishing referring expressions in long videos. We leave RL-based post-training to future work.

\end{revblock}

\noindent
\textbf{Future directions on Ref-SAV benchmarks.} As discussed in Tab.~\ref{tab:Ref-SAV-benchmark-comparison}, long text, occlusion inputs, and longer videos with camera motion are critical problems in our benchmarks. Thus, the new models need to design a more robust Ref-VOS system 
to handle these challenges with several specific designs, including more robust memory designs. In addition, understanding long text and improving long text grounding ability is also important to explore.

%% file: tables/t3_data_stat.tex
\begin{table}
    \centering
    \caption{Sa2VA training datasets.}
    \resizebox{1.\linewidth}{!}{
    \begin{tabular}{l|l}
    \toprule[0.2em]
    Type & Datasets \\
    \midrule
    Image QA & LLaVA 1.5 (665K)\\
    Image Seg & RefCOCO (17K), RefCOCO+ (17K), RefCOCOg (22K), Grand-f (214K)\\
    Video QA & Video-ChatGPT (100K)\\
    Video Seg & Ref-YTVOS (3.5K), MeVIS (0.6K), ReVOS (1.7K), \textbf{Ref-SAV (37K)}\\
    \bottomrule[0.1em]
    \end{tabular}
    }
    \label{tab:train_datasets}
\end{table}

%% file: tables/t_d1.tex
\begin{table}[t]
   \caption{\small Comparison with previous Ref-VOS benchmarks.}
    \resizebox{1.\linewidth}{!}{
   \begin{tabular}{r | c c c c c c}
      \toprule[0.15em]
     Property & DAVIS17-RVOS & ReVOS & Ref-YT-VOS & MeVIS & Ours  \\        \hline
         Short Text & \raisebox{-0.5ex}{\checkmark} & \raisebox{-0.5ex}{\checkmark} & \raisebox{-0.5ex}{\checkmark} & \raisebox{-0.5ex}{\checkmark} & \raisebox{-0.5ex}{\checkmark}\\
        Long Text & \raisebox{-0.5ex}{\ding{55}} & \raisebox{-0.5ex}{\checkmark}  & \raisebox{-0.5ex}{\ding{55}} & \raisebox{-0.5ex}{\ding{55}} & \raisebox{-0.5ex}{\checkmark}  \\
       Large Object Motion &  \raisebox{-0.5ex}{\ding{55}} & \raisebox{-0.5ex}{\ding{55}}  & \raisebox{-0.5ex}{\checkmark} & \raisebox{-0.5ex}{\ding{55}} & \raisebox{-0.5ex}{\checkmark}  \\
       Large Camera Motion &  \raisebox{-0.5ex}{\checkmark} & \raisebox{-0.5ex}{\checkmark} & \raisebox{-0.5ex}{\ding{55}} & \raisebox{-0.5ex}{\ding{55}} & \raisebox{-0.5ex}{\checkmark} \\
        Heavy Occlusion  &  \raisebox{-0.5ex}{\ding{55}} & \raisebox{-0.5ex}{\ding{55}}  & \raisebox{-0.5ex}{\ding{55}} & \raisebox{-0.5ex}{\ding{55}} & \raisebox{-0.5ex}{\checkmark} \\
      \bottomrule
   \end{tabular}
   }
   \label{tab:Ref-SAV-benchmark-comparison}
\end{table}

%% file: tables/t_d2_stat.tex
\begin{table}[t]
    \scriptsize
    \centering
   \vspace{-3mm}
    \caption{Statistics comparison with other Ref-VOS datasets.}
        \resizebox{1.\linewidth}{!}{
    \begin{tabular}{c|cccccc}
         \toprule[0.15em]
    Datasets & Video & Object & Expression & Mask & Avg. Len.  \\
    \midrule
    Ref-YTVOS & 3,978 & 7,451 & 15,009 & 131k & 9.68 \\
    MeViS & 2,006 & 8,171 & 28,570 & 443k & 7.07\\
    ReVOS & 1,042 & 9,084  & 35,074 & 273k & 10.5 \\
    \midrule
    Ref-SAV & 37,311 & 72,509 & 72,509 & 6.0m & 83.6 \\
    \bottomrule[0.1em]
    \end{tabular}
    \label{tab:tab_2_com_rebuttal}
    }
\end{table}

%% file: tables/t_impl.tex
\begin{table}[t]
\centering
\caption{Training strategies for model components.}
\label{tab:component_training_str}
        \resizebox{1.\linewidth}{!}{
\begin{tabular}{lllll}
\toprule
 & MLLM & SAM-2 Encoder & SAM-2 Decoder & SAM-2 Memory \\
\midrule
Training Strategy & LoRA & Frozen & Finetune & Frozen \\
\bottomrule
\end{tabular}
}
\end{table}

%% file: tables/t2_main.tex
\begin{table*}[t!]
    \centering
    \caption{\small{Experiment results on various image/video benchmarks. We report cIoU for the image segmentation sets, J\&F for the video segmentation sets, and AP50 for the GCG benchmark. For MME dataset~\cite{fu2023mme}, A / B denotes the perception (A) and cognition (B) scores, respectively, while C(+) represents the total score (C = A + B). Although it is more common to report perception and cognition separately, we report the sum as in their original paper if the individual scores are missing.}}
    \resizebox{0.99\textwidth}{!}{
    \begin{tabular}{c|ccc|cccc|ccc|cc|c}
    \toprule[0.2em]
    \multirow{2}{*}{Method}  & \multicolumn{3}{c|}{Image Segmentation} & \multicolumn{4}{c|}{Video Segmentation} & \multicolumn{3}{c|}{Image Chat} &  \multicolumn{2}{c|}{Video Chat} & \multicolumn{1}{c}{GCG} \\
    ~  & \scriptsize{RefCOCO~\cite{kazemzadeh2014referitgame}} &  \scriptsize{RefCOCO+~\cite{kazemzadeh2014referitgame}} &  \scriptsize{RefCOCOg~\cite{yu2016modeling}} &  \scriptsize{MeViS~\cite{MeViS}} &  \scriptsize{Ref-DAVIS17~\cite{khoreva2019video}} &  \scriptsize{Ref-YTVOS~\cite{seo2020urvos}}& \scriptsize{ReVOS~\cite{yan2024visa}} &  \scriptsize{MME~\cite{fu2023mme}} &  \scriptsize{MMBench~\cite{liu2023mmbench}} &  \scriptsize{SEED-Bench~\cite{li2023seed}} &  \scriptsize{Video-MME~\cite{fu2024video}} &  \scriptsize{MMBench-Video~\cite{fang2024mmbench}} &  \scriptsize{GCG~\cite{hanoona2023GLaMM}}\\
    \midrule
    LLAVA-1.5-13B~\cite{liu2023improvedllava} & - & - & - & - & - & - & - & 1531(+) & 68.8 & 70.1 & - & - & - \\
    Video-LLaVA-7B~\cite{lin2023videollava}  & - & - & - & - & - & - & - & - & 60.9 & - & 39.9 & 1.03 & -\\
    LLaMA-VID-7B~\cite{li2023llamavid} & - & - & - & - & - & - & - & 1521(+) & 65.1 & 59.9 & - & 1.08 & - \\
    mPLUG-Owl3-8B~\cite{ye2024mplug} & - & - & - & - & - & - & - & - & 77.6 & - & 53.5 & 1.35 & -\\
    InternVL2-8B~\cite{chen2024far} & - & - & - & - & - & - & - & - &81.7 & \textbf{76.2} & \textbf{54.0} & 1.28 & - \\
    PixelLM-7B~\cite{ren2023pixellm} & 73.0 & 66.3 & 69.3 & - & - & - & - & 309/135 & 17.4 & - &- &- & - \\
    LaSagnA~\cite{wei2024lasagna} & 76.8 & 66.4 & 70.6 & - & - & - & - & 0/0 & 0.0 & - & - & - & - \\
    LISA-7B~\cite{lai2023lisa} &74.9 & 65.1 & 67.9 & - & - & - & - & 1/1 & 0.4 & - &- & - & -\\
    GLaMM-7B~\cite{hanoona2023GLaMM} & 79.5 & 72.6 & 74.2 & - & - & - & - & 14/9 & 36.8 & - & - & - &28.9\\
    LLaVA-G-7B~\cite{zhang2025llavagrounding} & 77.1 & 68.8 & 71.5 & - & - & - & - & - & - & - & - & - & - \\
    GSVA-13B~\cite{xia2023gsva} & 79.2  & 70.3  & 75.7 & - & - & - & - & - & - & - & - & - & -\\
    OMG-LLaVA-7B~\cite{OMGLLaVA} & 78.0 & 69.1 & 72.9 & - & - & - & - & 1177/235 & 47.9 & 56.5& - & - & 29.9\\
    VideoLISA-3.8B~\cite{bai2024one} & 73.8 & 63.4 & 68.3 & 44.4 & 68.8 & 63.7 & - & - & - & - &- & - & -\\
    VISA-13B~\cite{yan2024visa} & 72.4 & 59.8 & 65.5 & 44.5 & 70.4 & 63.0 &  50.9 & - & -  & - &- & - & -\\
    \midrule
    Sa2VA-1B (Ours) & 77.4 & 69.9 & 72.3 & 41.7 & 72.3 & 65.3 &  47.6  & 1381/405 & 68.3 & 64.8 & 39.9 & 1.07 & 23.8 \\
    Sa2VA-4B (Ours) & 80.4 & 74.3 & 76.7 & 46.2 & 73.8 & 70.0 & 53.2 & 1553/540 & 76.8 & 72.6 & 50.4 & 1.23 & 28.2 \\
    Sa2VA-8B (Ours) & \textbf{81.9} & \textbf{76.5} & \textbf{78.9} & \textbf{46.9} & \textbf{75.2} & \textbf{70.7} & \textbf{57.6} & \textbf{1651/578} & \textbf{82.4} & 75.5 & 52.1 & \textbf{1.34} & \textbf{31.0} \\
    \midrule
    Sa2VA-26B (Ours) & 82.5 & 78.8 & 79.7 & 46.2 & 77.0 & 70.1 & 58.4 & 1691/538 & 83.7 & 76.8 &  52.6 & 1.45 & 33.5\\
    \bottomrule[0.1em]
    \end{tabular}
    }
    \label{tab:exp_01_ref_comparision}
\end{table*}

%% file: tables/t5_image_level_bench_twoc.tex
\begin{table*}[t!]
    \centering
    \caption{\small{Performance on image-level benchmarks with MLLMs that have segmentation capability. We report cIoU for the image segmentation sets. A / B denotes the perception (A) and cognition (B) scores, respectively, while C(+) represents the total score (C = A + B).}}
    \resizebox{1.\textwidth}{!}{
    \begin{tabular}{c|ccccccc|ccc}
    \toprule[0.2em]
    Method & MME~\cite{fu2023mme} & MMBench~\cite{liu2023mmbench} & SEED-Bench~\cite{li2023seed} & AI2D~\cite{kembhavi2016diagram} & MMStar~\cite{chen2024we} & MMMU~\cite{yue2023mmmu} & SQA$^{\rm test}$~\cite{lu2022sqa} & RefCOCO & RefCOCO+ & RefCOCOg\\
    \midrule
    LLAVA-1.5-13B~\cite{liu2023improvedllava} & 1531(+) & 68.8 & 70.1 & - & - & - & - & 0.0 & 0.0 & 0.0\\
    \midrule
     LISA-7B~\cite{lai2023lisa} & 1/1 & 0.4 & - & 0.0 & - & - & - & 74.9 & 65.1 & 67.9 \\
     PixelLM-7B~\cite{ren2023pixellm} & 309/135 & 17.4 & - &  0.0 & - & - & - & 73.0 & 66.3  & 69.3 \\
     LaSagnA-7B~\cite{wei2024lasagna} & 0/0 & 0.0 & - &  0.0 &  - & - & - & 76.8 &  66.4 & 70.6 \\
     GLaMM-7B~\cite{hanoona2023GLaMM} & 14/9 & 36.8 & - & 28.2 & - & - & - & 79.5 & 72.6 & 74.2  \\
     OMG-LLaVA-7B~\cite{OMGLLaVA} & 1177/235 & 47.9 & 56.5 & 42.9 & - & - & - & 78.0 & 69.1 & 72.9 \\
     \midrule
    Sa2VA-4B (ours) & 1553/540 & 76.8 & 72.6 & 79.9 & 53.7 & 46.2 & 95.8 & 80.4  & 74.3 & 76.7 \\
    Sa2VA-8B (ours) & 1651/578 &  82.4 & 75.5 & 82.1 & 60.3 & 44.7  & 96.8 & 81.9 & 76.5 & 78.9 \\
    \bottomrule[0.1em]
    \end{tabular}
    }
    \label{tab:image_benchmarks}
\end{table*}

%% file: tables/t8_intern25.tex
\begin{table*}[t!]
 \centering
\caption{\small{Experiment results using stronger InternVL2.5 in our Sa2VA.}}
\setlength{\tabcolsep}{2pt}
\resizebox{1.\textwidth}{!}{
\begin{tabular}{c|ccc|ccc|cccccc}
\toprule[0.2em]
Base & \multicolumn{3}{c|}{Image Segmentation} & \multicolumn{3}{c|}{Video Segmentation} & \multicolumn{6}{c}{Image Chat} \\
MLLM & \scriptsize{RefCOCO} & \scriptsize{RefCOCO+} & \scriptsize{RefCOCOg} & \scriptsize{MeViS (val\_u)} & \scriptsize{Ref-YTVOS} & \scriptsize{Ref-DAVIS17} & \scriptsize{MME} & \scriptsize{MMBench} & \scriptsize{SEED-Bench} &
\scriptsize{AI2D} & \scriptsize{MMStar} &
\scriptsize{SQA$^{\rm test}$} \\
\midrule
InternVL2.0-4B & 80.4 & 74.3 & 76.7 & 52.1 & 70.0 & 73.8 & 1553/540 & 76.8 & 72.6 & 79.9 & 53.7 & 95.8\\
InternVL2.0-8B & 81.9 & 76.5 & 78.9 & 57.0 & 70.7 & 75.2 & 1651/578 & 82.4 & 75.5 & 82.1 & 60.3 & 96.8 \\
\hline
InternVL2.5-1B & 79.6 & 73.6 & 77.7 & 53.4 & 68.0 & 69.5 & 1504/434 & 71.9 & 71.0 & 69.2 & 48.6 & 89.9 \\
InternVL2.5-4B & 82.4 & 77.6 & 79.7 & 55.9 & 71.3 & 73.7 & 1691/610 & 81.8 & 74.9 & 81.4 & 57.9 & 96.8 \\
InternVL2.5-8B & 82.6 & 78.0 & 80.3 & 58.9 & 72.3 & 75.9 & 1690/610 & 84.4 & 76.5 & 82.7 & 62.4 & 97.4\\
InternVL2.5-26B & 82.9 & 79.3 & 81.2 & 61.8 & 75.1 & 78.6 & 1698/653 & 85.8 & 78.3 & 85.7 & 67.0 & 98.4 \\
\bottomrule[0.1em]
\end{tabular}
}
\label{tab:exp_InternVL2_5}
\end{table*}

%% file: tables/t7_cotrain.tex
\begin{table*}[t]
    \centering
    \caption{\small{Ablation study on co-training effect on multiple datasets. We use Sa2VA-1B to test the performance.}}
    \resizebox{1.\textwidth}{!}{
    \begin{tabular}{c|ccc|cc|ccc|cc}
    \toprule[0.2em]
    \multirow{2}{*}{Data}  & \multicolumn{3}{c|}{Image Segmentation} & \multicolumn{2}{c|}{Video Segmentation} & \multicolumn{3}{c|}{Image Chat} &  \multicolumn{2}{c}{Video Chat} \\
    ~  & \scriptsize{RefCOCO} &  \scriptsize{RefCOCO+} &  \scriptsize{RefCOCOg} &  \scriptsize{MeViS (val\_u)} &  \scriptsize{Ref-DAVIS17} &   \scriptsize{MME} &  \scriptsize{MMBench} &  \scriptsize{SEED-Bench} &  \scriptsize{Video-MME} &  \scriptsize{MMBench-Video} \\
    \midrule
    All Data & 77.4 & 69.9 & 72.3 & 50.8 & 72.3 & 1381/405 & 68.3 & 64.8 & 39.9 & 1.07 \\
    \midrule
    w/o Image QA & 78.0 & 70.1 & 72.2 & 48.3 & 73.0 & 1298/359 & 63.4 & 63.8 & 39.7 & 0.39\\
    w/o Image Segmentation & 20.2 & 20.6 & 23.2 & 38.0 & 48.8 & 1393/408 & 70.1 & 65.7 & 41.2 & 1.08\\
    w/o Video QA & 78.0 & 70.4 & 72.6 & 50.7 & 74.3  & 1370/402 & 69.1 & 65.0 & 41.3 & 0.71 \\
    w/o Video Segmentation & 77.4 & 69.1 & 72.4 & 44.4 & 69.0  &1403/398& 67.8 & 64.9 & 40.4 & 1.04 \\
    \bottomrule[0.1em]
    \end{tabular}
    }
    \label{tab:data_ablation}
\end{table*}

%% file: tables/t4_refsav_results.tex
\begin{table}
    \centering
    \caption{\small{Ref-SAV validation sets. zs: zero-shot testing.}}
    \resizebox{1.\linewidth}{!}{
    \begin{tabular}{l|ccc| ccc | ccc}
    \toprule[0.2em]
    \multirow{2}{*}{Method}  & \multicolumn{3}{c|}{Long} & \multicolumn{3}{c|}{Short} & \multicolumn{3}{c}{Overall} \\
    ~ & J & F & J\&F & J & F & J\&F &J & F & J\&F \\
    \midrule
    UniRef++~\cite{wu2023uniref++} (zs) & 14.1 & 10.8 & 12.5 & 9.0 & 8.2 & 8.6 & 11.6 & 9.5 & 10.5\\
    UNINEXT~\cite{UNINEXT} (zs) & 11.7 & 8.3 & 10.0 & 5.8 & 4.4 &  5.1 & 8.8 & 6.4 & 7.6 \\
    MeVIS~\cite{MeViS} (zs) & 12.1 & 7.1 & 11.3 & 6.2 & 5.3 & 5.5 & 12.2 & 9.8 & 10.3  \\
    VISA~\cite{yan2024visa} (zs) & 16.1 & 12.2 & 14.1 & 12.3 & 9.6 & 9.2 & 13.2 & 11.3 & 11.8 \\
    \midrule
    Sa2VA-8b (zs) & 47.7 & 50.9 & 49.3 & 31.5 & 35.0 & 33.3 & 39.6 & 43.0 & 41.3 \\
    Sa2VA-8b (Ours) & 57.0 & 60.4 & 58.7 & 39.5 & 42.9 & 41.2 & 48.3 & 51.7 & 50.0 \\
    \bottomrule[0.1em]
    \end{tabular}
    }
    \label{tab:ref-sam-v-results}
\end{table}

%% file: tables/t6_finetune.tex
\begin{table}[t]
    \centering
    \caption{\small{Comparison with Fine-tuned Models.}}
    \resizebox{.85\linewidth}{!}{
    \begin{tabular}{c|ccc}
    \toprule[0.2em]
   Model Type & RefCOCO & RefCOCO+ & RefCOCOg \\
    \midrule
     LAVT~\cite{LAVT} &  72.7 & 62.1 & 61.2 \\
    GlaMM-7B~\cite{hanoona2023GLaMM} & 79.5 & 72.6 & 74.2 \\
    OMG-LLaVA-7B~\cite{OMGLLaVA} &  78.0 & 69.1 & 72.9 \\
    F-LLM-7B~\cite{wu2024flmm} & 76.1 & 65.2 & 68.5  \\
    \midrule
    Sa2VA-8B (ours) & 81.9 & 76.5 & 78.9\\
    Sa2VA-8B (ft) & 82.3 & 77.3 & 79.3 \\
    \bottomrule[0.1em]
    \end{tabular}
    }
    \label{tab:ft-ref-seg-results}
\end{table}

%% file: tables/t9_seg_token.tex
\begin{table}
    \centering
    \caption{\small{Ablation study on ``[SEG]'' token design.}}
    \resizebox{1.\linewidth}{!}{
    \begin{tabular}{c|ccc| cc}
    \toprule[0.2em]
    Type & RefCOCO & RefCOCO+ & RefCOCOg & DAVIS & MeVIS(val\_u)  \\
    \midrule
    Single & 77.4 & 69.9 & 72.3 & 72.3 & 50.8 \\
    Repetitive & 77.3 & 70.2 & 72.5 & 71.1 & 49.6 \\
    Unique & 77.6 & 70.3 & 72.4 & 68.6 & 46.3 \\
    \bottomrule[0.1em]
    \end{tabular}
    }
    \label{tab:seg_token}
\end{table}

%% file: tables/t10_dataset_ablation.tex
\begin{table}
    \centering
    \caption{\small{Ablation study on using more datasets.}}
    \resizebox{1.\linewidth}{!}{
    \begin{tabular}{c|c |cc| cc | c}
    \toprule[0.2em]
    Dataset & Size & RefCOCO & RefCOCOg & MMBench & MME & MeVIS(val\_u) 
    \\
    \midrule
    baseline & 1.2M & 77.4 & 72.3 & 68.3 & 1381/405 & 50.8 \\
    \midrule
    Inifinity-MM~\cite{gu2024infinity} &1.2M+3M & 77.1(\textcolor{red}{-0.3})& 72.6(\textcolor{green}{+0.3}) & 70.4(\textcolor{green}{+2.1}) & 1396/346(\textcolor{red}{-44}) & 51.2(\textcolor{green}{+0.4}) \\
    Ref-SAV &1.2M+37K & 77.2(\textcolor{red}{-0.2}) & 72.6(\textcolor{green}{+0.3}) & 68.2(\textcolor{red}{-0.1}) & 1384/418(\textcolor{green}{+16}) & 52.5(\textcolor{green}{+1.7}) \\
    \bottomrule[0.1em]
    \end{tabular}
    }
    \label{tab:scale_data}
\end{table}

%% file: tables/qwen_table.tex
\begin{table*}[t!]
\centering
\caption{\small{Experiment results of Sa2VA with recent base MLLMs.}}
\resizebox{1.\textwidth}{!}{
\begin{tabular}{c|ccc|ccc|cccccc}
\toprule[0.2em]
Base & \multicolumn{3}{c|}{Image Segmentation} & \multicolumn{3}{c|}{Video Segmentation} & \multicolumn{6}{c}{Image Chat} \\
MLLM & \tiny{RefCOCO} & \tiny{refCOCO+} & \tiny{refCOCOg} & \tiny{MeViS (val\_u)} & \tiny{ReVOS} & \tiny{Ref-DAVIS17} & \tiny{MME} & \tiny{MMBench} & \tiny{SEED-Bench} &
\tiny{AI2D} & \tiny{MMStar} &
\tiny{SQA$^{\rm test}$} \\
\midrule
Sa2VA-8B & 81.9 & 76.5 & 78.9 & 57.0 & 57.6& 75.2 & 1651/578 & 82.4 & 75.5 & 82.1 & 60.3 & 96.8 \\
\midrule
Sa2VA-InternVL3-2B & 81.4 & 75.7 & 80.3 & 53.9 & 56.2 & 74.5 & 1631/559 & 79.8 & 73.9 & 77.1 & 59.2 & 93.7 \\
Sa2VA-InternVL3-8B & 83.3 & 78.9 & 81.8 & 56.4 & 60.8 & 76.3 & 1743/633 & 83.0 & 76.2 & 84.3 & 65.9 & 97.5 \\
Sa2VA-InternVL3-14B & 83.6 & 79.9 & 83.6 & 59.2 & 60.7 & 76.6 & 1746/724 & 84.3 & 76.6 & 85.2 & 67.4 & 98.7 \\
\hline
Sa2VA-Qwen2.5VL-3B & 79.6 & 74.0 & 77.1 & 51.6 & 52.0 & 73.4 & 1533/572 & 78.4 & 73.9 & 81.1 & 57.7 & 80.3 \\
Sa2VA-Qwen2.5VL-7B & 82.4 & 77.5 & 81.5 & 56.4 & 58.3 & 79.4 & 1552/676 & 84.5 & 75.0 & 84.5 & 62.3 & 87.9 \\
Sa2VA-Qwen3VL-4B & 81.7 & 77.4 & 80.0 & 57.1 & 56.7 & 75.9 & 1660/655 & 86.3 & 77.3 & 85.4 & 66.3 & 91.6 \\
\bottomrule[0.1em]
\end{tabular}
}
\label{tab:exp_Sa2VA_Qwen}
\end{table*}

%% file: tables/t_a1_comp.tex
\begin{table}[t]
    \centering
    \caption{\small{Comparison with Recent Video MLLMs.}}
    \resizebox{.85\linewidth}{!}{
    \begin{tabular}{c|ccc}
    \toprule[0.2em]
   Model Type & MeViS & ReVOS & Ref-DAVIS17 \\
    \midrule
    PG-Video-LLaVA~\cite{munasinghe2023pg} & 18.9 & - & - \\
    GLaMM + SAM2~\cite{munasinghe2024videoglamm} & 38.7 & - & - \\
     VideoGLaMM (3.8B)~\cite{munasinghe2024videoglamm} & 45.2 & - & - \\
     VISA-13B~\cite{yan2024visa} & 44.5 & 50.9 & 70.4 \\
     VideoLISA-3.8B~\cite{bai2024one} & 44.4 &  - & 68.8\\
     HyperSeg-3B~\cite{wei2024hyperseg}& -& 55.7 & 71.2\\
     InstructSeg~\cite{wei2024instructseg} & - & 54.5 & 71.1\\
    \midrule
    Sa2VA-4B (ours) & 46.2 & 53.2 & 73.8  \\
    Sa2VA-8B (ours) & 46.9 & 57.6 & 75.2\\
    \bottomrule[0.1em]
    \end{tabular}
    }
    \label{tab:video_mllm_models}
\end{table}

%% file: tables/t_a2_comp_expert.tex
\begin{table}[t]
    \centering
    \caption{\small{Comparison with Vision Expert Models.}}
    \resizebox{1.\linewidth}{!}{
    \begin{tabular}{c|ccc|cc}
    \toprule[0.2em]
   Model Type & RefCOCO & RefCOCO+ & RefCOCOg & MeVIS & Ref-DAVIS17 \\
    \midrule
    LAVT~\cite{LAVT} &  72.7 & 62.1 & 61.2 & - & - \\
    ReferFormer~\cite{wu2022language} & - & - & - & 31.0 & 61.1 \\
    UniRef++-L~\cite{wu2023uniref++} &  81.4 & 74.0 & 76.0 & - & 67.2 \\
    EVF-SAM~\cite{zhang2024evfsam} & 82.4 & 76.5 & 78.2 & - & - \\
    LMPM~\cite{MeViS} & - & - & - & 37.2 & - \\
    UniVS~\cite{li2024univs} & - & - & - & - &59.4\\
    \midrule
    Sa2VA-26B (ours) &  82.5 & 78.8 & 79.7 &  46.2 & 77.0 \\
    \bottomrule[0.1em]
    \end{tabular}
    }
    \label{tab:vision_expert_model}
\end{table}

%% file: tables/t_a3_refsav_training_effect.tex
\begin{table}[t]
    \centering
    \caption{Vision Expert on Ref-SAV validation sets.}
    \resizebox{1.\linewidth}{!}{
    \begin{tabular}{l|ccc| ccc | ccc}
    \toprule[0.2em]
    \multirow{2}{*}{Method}  & \multicolumn{3}{c|}{Long} & \multicolumn{3}{c|}{Short} & \multicolumn{3}{c}{Overall} \\
    ~ & J & F & J\&F & J & F & J\&F &J & F & J\&F \\
    \midrule
    UniRef++~\cite{wu2023uniref++} (zero-shot) & 14.1 & 10.8 & 12.5 & 9.0 & 8.2 & 8.6 & 11.6 & 9.5 & 10.5\\
    \midrule
    UniRef++ (fine-tuning) & 19.2 & 15.1 & 17.2 & 12.3 & 11.7 & 12.0 & 15.8 & 13.4 & 14.6 \\
    \bottomrule[0.1em]
    \end{tabular}
    }
    \label{tab:effectiveness_on_more_baselines}
\end{table}

%% file: tables/t_a4_region_caption.tex
\begin{table}[t]
    \centering
    \caption{\small{Region caption performance on RefCOCOg dataset.}}
    \resizebox{1.\linewidth}{!}{
    \begin{tabular}{l|c|ccccc}
    \toprule[0.2em]
    Method & Sa2VA-4B & OMG-LLaVA~\cite{OMGLLaVA} & Osprey~\cite{yuan2023osprey} & GLaMM~\cite{hanoona2023GLaMM} & 
    GRIT~\cite{wu2025grit} & Kosmos-2~\cite{peng2023kosmos}\\
    \hline
    METEOR &  17.3 & 15.3 & 16.6 & 16.2 & 15.2 & 14.1 \\
    \bottomrule[0.1em]
    \end{tabular}
    }
    \label{tab:effectiveness_visual_prompt_tasks}
\end{table}

%% file: tables/t_ana_1.tex
\begin{table}[t]
\centering
\caption{Computational cost comparision.}
\label{tab:benchmark}
\resizebox{1.\linewidth}{!}{
\begin{tabular}{l c l l l c}
\toprule
\textbf{Model} & \textbf{Params} & \textbf{Base MLLM} & \textbf{Base Language Model} & \textbf{GPU} & \textbf{Inference Time} \\
\midrule
Sa2VA-1B  & 1B  & InternVL2.5-1B  & Qwen/Qwen2.5-0.5B-Instruct  & NVIDIA-H100 & 0.123s \\
Sa2VA-4B  & 4B  & InternVL2.5-4B  & Qwen/Qwen2.5-3B-Instruct  & NVIDIA-H100 & 0.282s \\
Sa2VA-8B  & 8B  & InternVL2.5-8B  & internlm/internlm2\_5-7b-chat & NVIDIA-H100 & 0.201s \\
Sa2VA-26B & 26B & InternVL2.5-26B & internlm/internlm2\_5-20b-chat & NVIDIA-H100 & 0.463s \\
\bottomrule
\end{tabular}
}
\end{table}

%% file: tables/t_ana_2.tex
\begin{table}[t]
\centering
\caption{Ablation study on keyframe sampling strategies using Sa2VA-8B on the MeVIS (val\_u) dataset.}
\label{tab:sampling_strategy}
\resizebox{1.\linewidth}{!}{
\begin{tabular}{l c c c}
\toprule
\textbf{Sampling Strategy} & \textbf{MeVIS J\&F (val\_u)} & \textbf{Inference Time (per sample)} & \textbf{\# Image Tokens} \\
\midrule
First 1 Frame & 55.1 & 0.151s & 256 \\
First 3 Frames & 58.7 & 0.167s & 768 \\
First 4 Frames & 59.5 & 0.181s & 1024 \\
First 5 Frames & 58.9 & 0.207s & 1280 \\
\textbf{Uniform 5 Frames} & \textbf{62.9} & \textbf{0.207s} & \textbf{1280} \\
\bottomrule
\end{tabular}}
\end{table}

%% file: tex/5_con.tex
\section{Conclusion}
\label{sec:conclusion}
In this work, we present Sa2VA, a versatile framework that integrates SAM-2 with MLLMs to achieve a dense, grounded understanding of both images and video. 
Our method can handle various image and video understanding tasks, including referring image/video segmentation and image/video conversation, with just single-stage instruction tuning.
By leveraging the knowledge from both MLLM and SAM-2, our model has strong capabilities in both mask and language generation. 
To demonstrate the effectiveness of our proposed method, we propose a challenging referring video object segmentation benchmark, Ref-SAV.
Extensive experimental results show that Sa2VA achieves strong performance on various tasks across benchmarks.
In addition, Sa2VA can also be extended with various modern MLLMs, with potential for building stronger baselines for pixel multi-modal systems.

\section*{Acknowledgments}
Supported by the Intelligence Advanced Research Projects Activity (IARPA) via Department of Interior / Interior Business Center (DOI/IBC) contract number 140D0423C0074. The U.S. Government is authorized to reproduce and distribute reprints for Governmental purposes notwithstanding any copyright annotation thereon. Disclaimer: The views and conclusions contained herein are those of the authors and should not be interpreted as necessarily representing the official policies or endorsements, either expressed or implied, of IARPA, DOI/IBC, or the U.S. Government. This work was also supported in part by the Institute of Information \& Communications Technology Planning \& Evaluation (IITP) grant funded by the Korean Government (MSIT) (No. RS-2024-00457882, National AI Research Lab Project).

%% file: tex/6_appendix.tex
\section{Concrete Benefits of the Unified Framework}
\label{sec:appendix_unified}
\begin{revblock}
\noindent
Sa2VA employs a single set of weights for all tasks, rather than training a separate model for each input type and task. Beyond the summary in Sec~3.2, we detail four benefits of this unified design, each supported by the experiments in the main paper.

\noindent
\textbf{Positive cross-task transfer.} Joint co-training across the image/video~$\times$~chat/segmentation grid yields gains that single-task training cannot recover. As shown in the co-training ablation (Tab. 9), removing any single modality of training data causes large drops; for instance, removing image-segmentation data collapses RefCOCO from 77.4 to 20.2 cIoU, and removing video-segmentation data degrades MeViS and Ref-DAVIS17 by 6.4 and 3.3 J\&F. The unified paradigm is thus essential, independently of any newly introduced data.

\noindent
\textbf{No catastrophic forgetting of conversational ability.} A recurring failure mode of grounding MLLMs is that instruction tuning for segmentation destroys their chat ability. Sa2VA instead retains the conversational performance of its base MLLM across MME, MMBench, SEED-Bench, and AI2D while simultaneously reaching state-of-the-art grounding (Tab.~6, Tab.~7).

\noindent
\textbf{Lower maintenance than per-dataset fine-tuning.} Fine-tuning the model on each individual dataset yields only marginal improvements (Tab.~11, 1.5\% on the referring sets and only 0.4 on RefCOCO for single-dataset fine-tuning), yet it requires a separately stored and maintained checkpoint per dataset. The unified model attains comparable or better accuracy with a single checkpoint.

\noindent
\textbf{Deployment and inference efficiency.} A single set of weights serves all tasks with no per-task model switching cost. The appended SAM-2 adds only $\approx$220M parameters and runs at 39.5 FPS, so the overall computational cost is dominated by the shared auto-regressive MLLM (Tab.~19: 0.123\,s for Sa2VA-1B, 0.282\,s for 4B, 0.201\,s for 8B, and 0.463\,s for 26B). The visual-prompt / ``[SEG]'' token budget is identical across tasks, so unification does not introduce additional token overhead.
\end{revblock}

\section{Annotation Model Choice and Validation}
\label{sec:appendix_annotation}
\begin{revblock}
\noindent
\textbf{Choice of InternVL2-76B.} We adopted InternVL2-76B as the primary captioner because it was one of the strongest open-source MLLMs available when the Ref-SAV training set was built. Importantly, the annotation pipeline (Fig.~3, Sec.~3.3) is not a single-model pseudo-labeler: object-level captions generated by InternVL2-76B are cross-checked by Qwen2-72B, and inconsistent or unrecognizable cases are discarded. This two-model agreement filter reduces the risk of systematic single-model bias.

\noindent
\textbf{Decoupled annotation model and backbones.} Using an InternVL-family model for both annotation and the backbone does not introduce a family-internal advantage. First, the annotation model is InternVL2-76B, whereas the main experiments adopt backbones spanning InternVL2-1B/4B/8B/26B, InternVL2.5, InternVL3, and the unrelated Qwen2.5-VL and Qwen3-VL families (Tab.~8 and Tab.~14), which have distinct weights and even distinct model families. Second, Ref-SAV transfers cleanly to backbones outside the InternVL family: UniRef++, which is unrelated to InternVL, improves from 10.5 to 14.6 overall J\&F after fine-tuning on Ref-SAV (Tab.~17), and the Qwen-family backbones obtain similarly large gains (Tab.~14). Such cross-family transfer would not arise from family-internal label leakage.

\noindent
\textbf{Validation of annotation quality.} The Ref-SAV evaluation is anchored to human ground truth: the short-expression evaluation set is fully human-annotated, and the long-expression evaluation set is human-filtered (Sec.~3.3). Therefore all reported test numbers measure agreement with human labels rather than with the automatic captioner. As external validation, Tab.~17 shows that an independent method consistently improves after training on our automatic labels, across both the long (12.5$\rightarrow$17.2 J\&F) and short (8.6$\rightarrow$12.0 J\&F) splits, indicating that the automatic data engine produces useful supervision.
\end{revblock}

\begin{table*}[!t]
\centering
\caption{\rev{Ablation study on the segmenter choice under a fixed Sa2VA pipeline (Qwen3-VL-4B).}}
\label{tab:sam3_ablation}
\small
\setlength{\tabcolsep}{6pt}
\begin{tabular}{lcccccc}
\toprule
\multirow{2}{*}{Segmenter} & RefCOCO & RefCOCO+ & RefCOCOg & MeViS & ReVOS & Ref-DAVIS17 \\
 & (cIoU) & (cIoU) & (cIoU) & (J\&F) & (J\&F) & (J\&F) \\
\midrule
SAM-2 & 83.6 & 79.2 & 82.4 & 61.4 & 61.4 & 76.2 \\
SAM-3 & \textbf{83.7} & \textbf{79.4} & \textbf{83.0} & \textbf{65.3} & \textbf{66.3} & \textbf{77.1} \\
\bottomrule
\end{tabular}
\end{table*}

\begin{table*}[!t]
\centering
\caption{\rev{Controlled same-backbone comparison against LISA-7B. All Sa2VA variants share an identical training recipe.}}
\label{tab:backbone_fairness}
\small
\setlength{\tabcolsep}{6pt}
\begin{tabular}{lcccccc}
\toprule
\multirow{2}{*}{Model (backbone)} & RefCOCO & RefCOCO+ & RefCOCOg & MeViS & ReVOS & Ref-DAVIS17 \\
 & (cIoU) & (cIoU) & (cIoU) & (J\&F) & (J\&F) & (J\&F) \\
\midrule
LISA-7B (LLaVA-1.5-7B)$^{\dagger}$ & 74.9 & 65.1 & 67.9 & -- & -- & -- \\
\midrule
Sa2VA (LLaVA-1.5-7B) & 80.3 & 73.1 & 79.2 & 54.8 & 54.0 & 74.6 \\
Sa2VA (Qwen2.5-VL-7B) & 82.3 & 77.1 & 80.5 & 58.1 & 58.6 & \textbf{77.4} \\
Sa2VA (Qwen3-VL-4B) & \textbf{83.6} & \textbf{79.2} & \textbf{82.4} & \textbf{61.4} & \textbf{61.4} & 76.2 \\
\bottomrule
\end{tabular}
\end{table*}

\section{Generalization to the Newer Segmenter: SAM-3}
\label{sec:appendix_sam3}
\begin{revblock}
\noindent
Generalization to evolving segmentation models is an explicit design goal of Sa2VA. The decoupled design (Sec.~3.2) keeps the segmenter's encoder and memory frozen and bridges to the MLLM only through the spatial-temporal ``[SEG]'' prompt token, so any prompt-able segmentation decoder can be plugged in without modifying the MLLM side. To verify this, we replace SAM-2 with the more recent SAM-3 while keeping the rest of the pipeline unchanged, using a Qwen3VL-4B backbone. Since SAM-3 retains a promptable visual segmentation interface, the ``[SEG]'' token transfers directly through the existing prompt projection. To isolate the effect of the segmenter, both rows are trained under an identical recipe \emph{without} the Ref-SAV data, so that the comparison reflects the segmenter alone rather than any additional training data. We train Sa2VA with publicly released codebase and have released the model and training config for reproduce.

\noindent
As reported in Tab.~\ref{tab:sam3_ablation}, upgrading the segmenter consistently improves performance, with the largest gains on the video benchmarks (e.g., MeViS $61.4\rightarrow65.3$ and ReVOS $61.4\rightarrow66.3$ J\&F). This result supports two conclusions: (i) the framework is modular with respect to the segmenter, as the decoupled design and the ``[SEG]'' bridge enable a drop-in replacement; and (ii) the framework benefits from stronger segmenters without retraining the MLLM. We note that the contribution of Sa2VA lies not in adopting a stronger segmenter, but in unifying diverse tasks within a single architecture. The SAM-3 integration is also released in our public repository to facilitate further research.
\end{revblock}

\section{Backbone Fairness: Same-Backbone Comparison}
\label{sec:appendix_backbone}
\begin{revblock}
\noindent
To isolate the contribution of our unified paradigm from that of a stronger backbone, we drop a LISA-comparable language backbone---the original LLaVA-1.5-7B---into our framework and compare against LISA-7B under the same backbone. All Sa2VA rows in Tab.~\ref{tab:backbone_fairness} are trained from scratch under an identical recipe (1 epoch, LoRA $r=128$, frozen LLM and visual encoder; image scores are val cIoU and video scores are J\&F), \emph{without} the Ref-SAV data, so that the comparison reflects the backbone alone rather than any additional training data. On the same LLaVA-1.5-7B backbone, our paradigm lifts referring segmentation well above LISA-7B ($+5.4$ RefCOCO, $+8.0$ RefCOCO+, $+11.3$ RefCOCOg) and additionally endows it with referring \emph{video} segmentation that LISA does not support at all. The gain therefore stems from the unified paradigm rather than a stronger backbone; swapping in stronger MLLM backbones (Qwen2.5-VL, Qwen3-VL) improves the numbers further, but even the deliberately weak LLaVA-1.5 backbone already surpasses the prior specialist. We train Sa2VA with our publicly released codebase and have released the trained models and training configs for reproducibility.
\end{revblock}

\section{Disentangling Data and Paradigm}
\label{sec:appendix_data_paradigm}
\begin{revblock}
\noindent
A key question is whether the performance of Sa2VA stems from the newly constructed data or from the unified training paradigm. We first note that Sa2VA is efficient to train: it requires no pre-training stage and only a single epoch of supervised instruction tuning with LoRA ($r=256$), taking $\approx$24 hours on 16 NVIDIA H100 80GB GPUs (bf16, total batch size 256), as detailed in Sec.~4. We then disentangle the two factors from both directions.

\noindent
\textbf{The paradigm matters with data held fixed.} The co-training ablation (Tab.~9) is conducted \emph{without} the Ref-SAV data, and removing any single modality of training data already causes large drops. Hence the joint paradigm is necessary independently of the new dataset.

\noindent
\textbf{The same data cannot compensate for a weaker paradigm.} Fine-tuning a non-unified baseline (UniRef++) on the same Ref-SAV data improves its overall J\&F only from 10.5 to 14.6 (Tab.~17), far below Sa2VA. Under an identical data-construction cost, a conventional paradigm cannot approach the unified framework, indicating an inherent advantage of the framework rather than a mere data advantage.

\noindent
\textbf{Targeted data helps, but only through the paradigm.} With the paradigm fixed, adding 3M generic VQA samples barely affects segmentation, whereas adding only 37K targeted Ref-SAV samples yields $+1.7$ J\&F on MeViS (Tab.~13). The paradigm requires data targeted to the task, and the data is effective only within the paradigm. In summary, data and paradigm are complementary, and neither factor alone accounts for the performance of Sa2VA.
\end{revblock}

\section{The VQA--Grounding Representation Conflict}
\label{sec:appendix_vqa_grounding}
\begin{revblock}
\noindent
We expand on the hypothesis introduced in Sec.~4.4 regarding why scaling up VQA data can hurt referring segmentation.

\noindent
\textbf{Why VQA and grounding compete in the shared token space.} VQA is optimized with a next-token language-modeling loss whose gradient pressure favors holistic, sequence-level semantic representations: visual tokens are encouraged to be compressed into a compact summary suitable for autoregressive answer generation. Referring segmentation, in contrast, requires the hidden state of the ``[SEG]'' token to retain fine-grained, spatially discriminative information, so that the SAM-2 decoder can localize the correct pixels. As VQA data is scaled up, the representations of the LLM drift toward the global-semantic objective, weakening the spatial signal carried by the ``[SEG]'' token and degrading segmentation. This representation-objective conflict is most visible in Tab.~9: removing image-segmentation supervision collapses RefCOCO from 77.4 to 20.2 cIoU, showing how fragile the spatial signal carried by the ``[SEG]'' token is once the training objective shifts away from dense grounding. Consistent with this, scaling generic VQA data (Tab.~13) yields no segmentation improvement even as it raises chat scores.

\noindent
\textbf{Visual generative tasks may align better with dense grounding.} Generative objectives such as image/video synthesis or masked reconstruction require spatially grounded representations for pixel-level reconstruction, which are structurally closer to dense grounding than the holistic-semantic objective of VQA. We thus conjecture that jointly training with generative objectives may interfere less with grounding than VQA does, and could even act as a mutual regularizer. Validating this empirically, for example by adding a masked image/video reconstruction head or a diffusion-style auxiliary loss, is a promising direction for future work.

\noindent
\textbf{Possible mitigations within the current paradigm.} Beyond generative co-training, we are exploring (i) task-specific token subspaces with lightweight gating, (ii) representation-level regularization that constrains ``[SEG]'' features to a separable subspace, and (iii) reinforcement learning with verifiable segmentation rewards (Sec.~5).
\end{revblock}

\section{Additional Qualitative Results}
\label{sec:appendix_visualization}
\begin{revblock}
\noindent
Owing to the page limit, the following qualitative results on video referring segmentation, visual prompt understanding, and failure cases are presented here rather than in the main text.
\end{revblock}

\begin{figure*}[!t]
  \centering
  \includegraphics[width=0.85\linewidth]{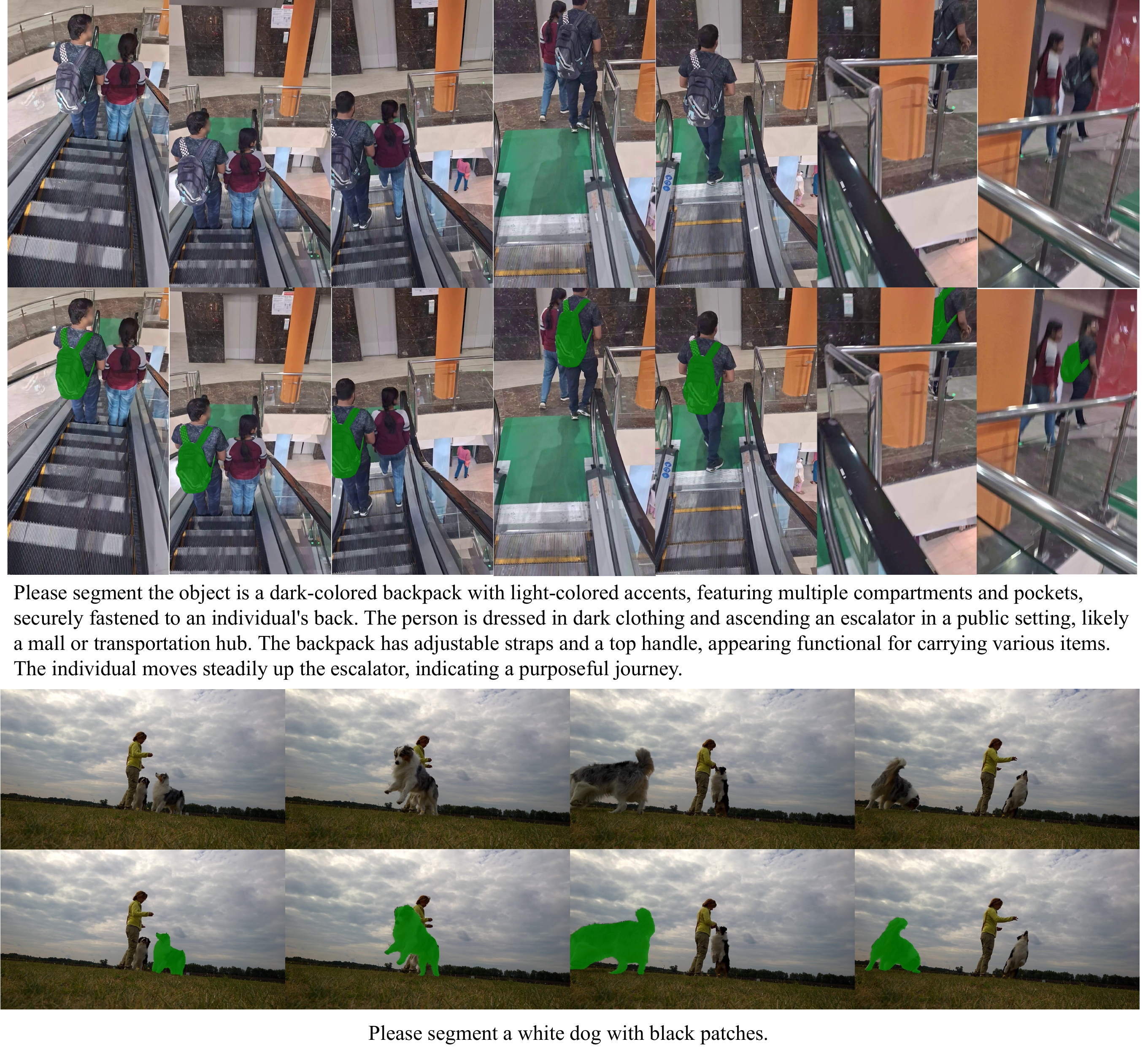}
  \caption{Visualization results on video referring segmentation.}
  \label{fig:vid_demo}
\end{figure*}
\begin{figure*}[!t]
  \centering
  \includegraphics[width=0.85\linewidth]{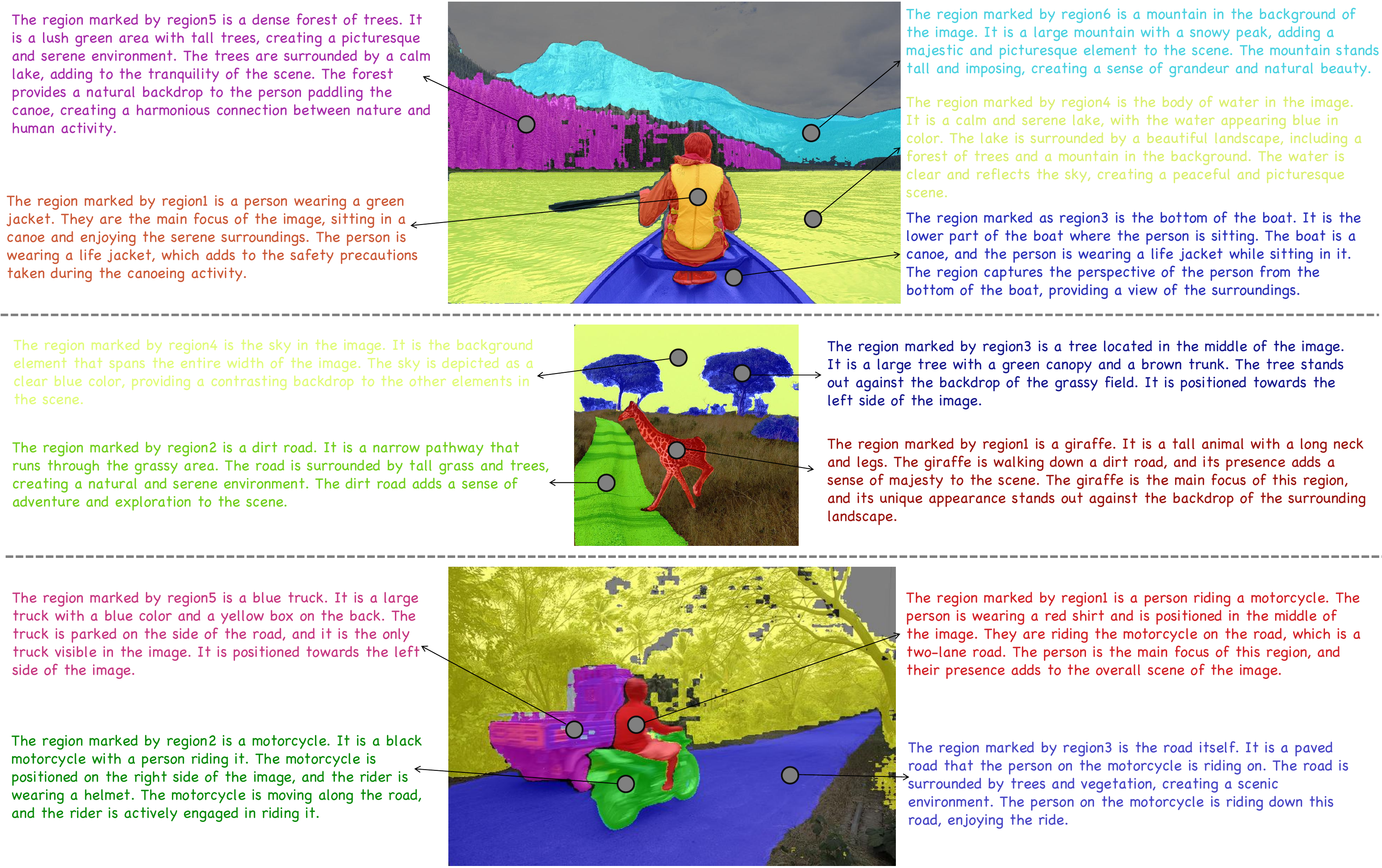}
  \caption{Visualization results on the visual prompt understanding task. We use the masks predicted by our model under the GCG task as visual prompts, and generate region-level descriptions for these masks. The object masks and their corresponding region captions are highlighted in the same color.}
  \label{fig:vp_demo}
\end{figure*}
\begin{figure*}[!t]
  \centering
  \includegraphics[width=0.85\linewidth]{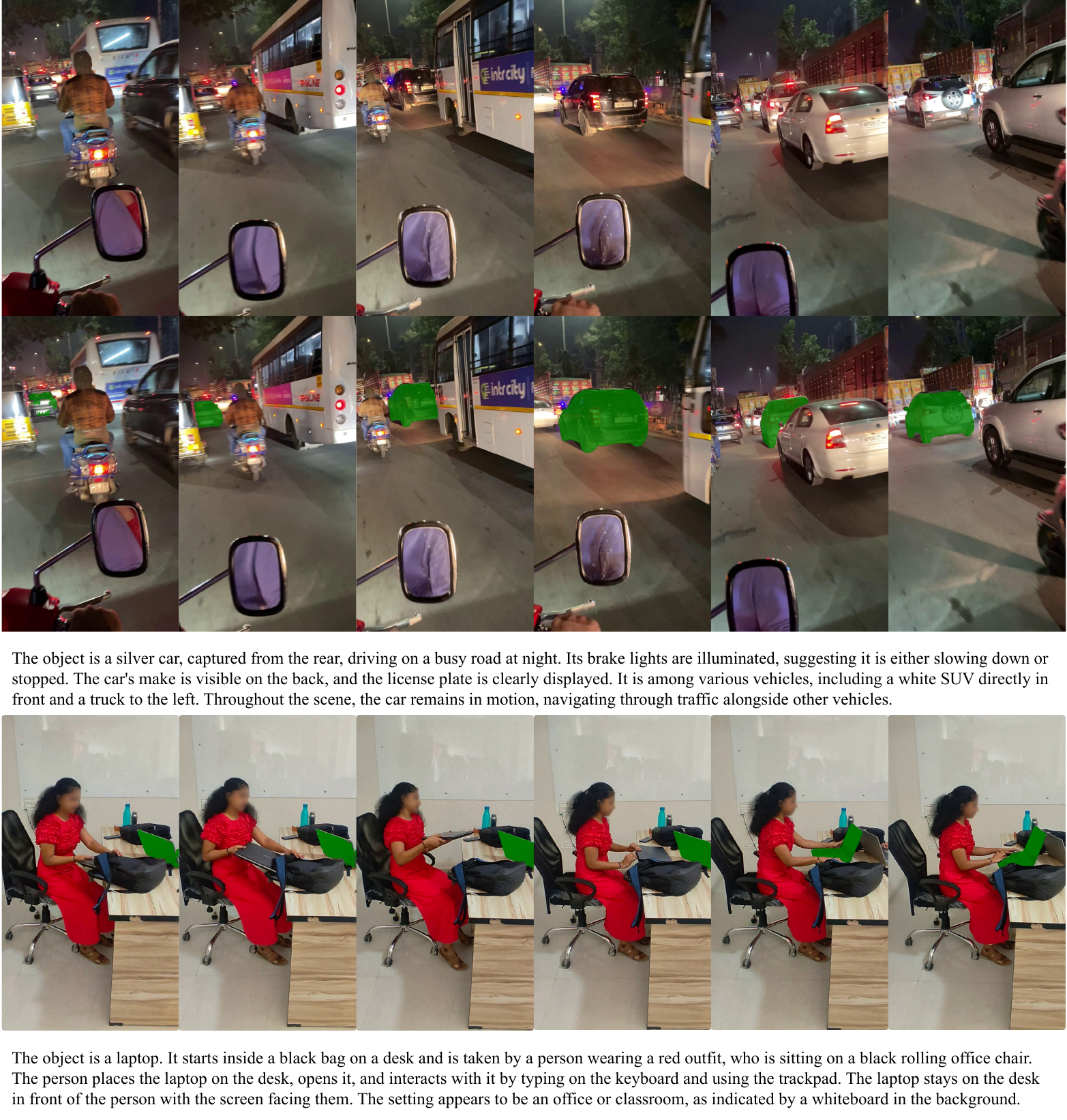}
  \caption{Visualization failure cases.}
  \label{fig:fail_demo}
\end{figure*}

\begin{revblock}
\noindent
\textbf{Results on video referring segmentation.}
As shown in Fig.~\ref{fig:vid_demo}, our method performs well in handling diverse and challenging conditions, showcasing a remarkable ability to adjust and perform effectively even in occlusion scenes or highly dynamic environments.

\noindent
\textbf{Results on visual prompt understanding.}
As shown in Fig.~\ref{fig:vp_demo}, our method can generate descriptions based on the visual prompts. The descriptions generated by our method demonstrate a high degree of contextual awareness, effectively capturing details within the visual cues.

\noindent
\textbf{Failure cases.}
As shown in Fig.~\ref{fig:fail_demo}, we present representative failure cases, which mainly arise from long videos with hard, fine-grained referring expressions, as discussed in Sec.~5.
\end{revblock}